%% file: main.tex
\pdfoutput=1
\documentclass[11pt]{article}

\usepackage[final]{acl}
\usepackage{ulem}
\usepackage{times}
\usepackage{latexsym}
\usepackage{xspace}
\usepackage{bm}

\usepackage{makecell}

\usepackage{amsmath}   % advanced math environments (equation, align, etc.)
\usepackage{amssymb}   % extra symbols
\usepackage{amsfonts}  % \mathbb, \mathbb{E}, …
\usepackage{multicol} 
\usepackage{multirow}
\usepackage{cutwin}
\usepackage{colortbl}
\input{preamble}
\usepackage{mathabx}
\usepackage{algorithm}
\usepackage{algorithmicx}
\usepackage{algpseudocode}
\usepackage[T1]{fontenc}
\usepackage[utf8]{inputenc}

\usepackage{microtype}
\usepackage{enumitem} 

\usepackage{inconsolata}

\usepackage{graphicx}
\usepackage{booktabs}
\newcommand{\method}{\textsc{RACCooN}\xspace}

\title{\method: Versatile Instructional Video Editing\\
with Auto-Generated Narratives}

\author{Jaehong Yoon$^{1,2,} $\thanks{Equal Contribution.}\quad\quad Shoubin Yu$^{1,}\footnotemark[1]$ \quad\quad Mohit Bansal$^{1}$\\
$^{1}$UNC Chapel Hill \quad\quad $^2$Nanyang Technological University\\
{\tt jaehong.yoon@ntu.edu.sg\quad\quad \{shoubin, mbansal\}@cs.unc.edu}
\\
\\
\url{https://raccoon-mllm-gen.github.io/}
}

\begin{document}
\maketitle
\input{sections/0_abs}
\input{sections/1_intro}
\input{sections/2_related}

\input{sections/3_method}
\input{sections/4_exp}

\input{sections/5_conclusion}

{
    % \small
    % \bibliographystyle{ieeenat_fullname}
    \bibliography{ref}
}

\clearpage
\appendix
\section*{\LARGE{Appendix}}                                   
\input{sections/6_appendix}

\end{document}

%% file: preamble.tex
\definecolor{darkgreen}{rgb}{0,0.5,0}
\definecolor{azureblue}{rgb}{0,0.5,1}
\definecolor{darkgreen}{rgb}{1,0,0}
\definecolor{color1}{HTML}{006EB8}
\definecolor{color2}{HTML}{009B55}
\definecolor{color3}{HTML}{924DBF}
\definecolor{BrickRed}{rgb}{0.8, 0.25, 0.33} % 약간 어두운 빨강
\definecolor{Green}{rgb}{0.0, 0.45, 0.2}

\usepackage{pifont}

\hypersetup{
    colorlinks=true,
    linkcolor=magenta,
    filecolor=magenta,  
    urlcolor=magenta,
    citecolor=color3,
    pdftitle={Overleaf Example},
    pdfpagemode=FullScreen,
    }
\makeatletter
\renewcommand\@makefnmark{\hbox{\normalfont\textsuperscript{\textcolor{black}{\@thefnmark}}}}
\makeatother
\usepackage[capitalize]{cleveref}
\crefname{section}{Sec.}{Secs.}
\Crefname{section}{Section}{Sections}
\Crefname{table}{Table}{Tables}
\crefname{table}{Tab.}{Tabs.}
\crefname{appendix}{Sec.}{Secs.}
\Crefname{appendix}{Section}{Sections}

\definecolor{ggs}{gray}{0.45}
\definecolor{ggg}{gray}{0.65}
\definecolor{gg}{HTML}{F2DFC2}

%% file: sections/0_abs.tex
\begin{figure*}
    \centering
    \vspace{-0.2in}\includegraphics[width=\linewidth]{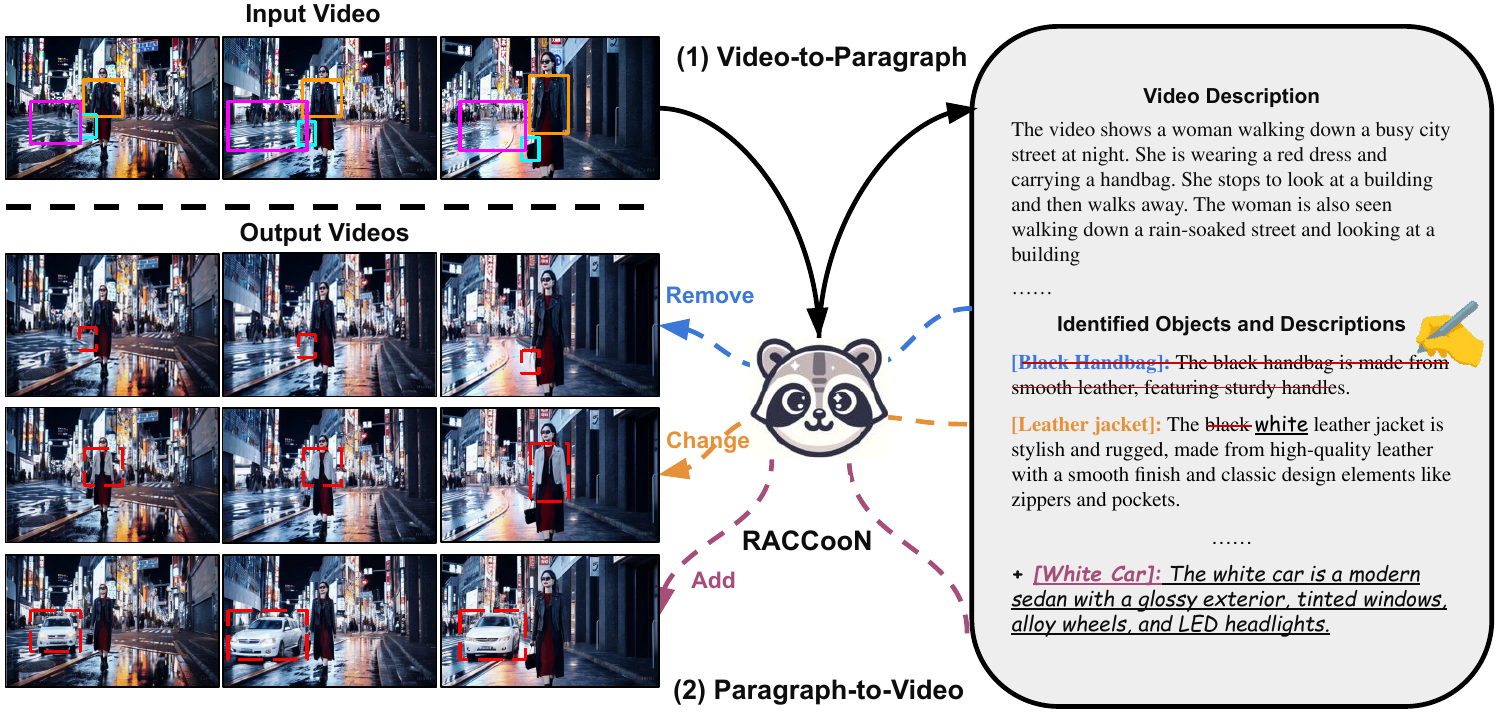}    
    \caption{
    \textbf{Overview of \method}, a versatile and user-friendly video-to-paragraph-to-video framework, enables users to remove, add, or change video content via updating auto-generated narratives. 
    }
    \label{fig:teaser}
\end{figure*}
% \jy{remove 'framework' from the title?}

\begin{abstract}
Recent video generative models primarily rely on detailed, labor-intensive text prompts for tasks, like inpainting or style editing, limiting adaptability for personal/raw videos. 
This paper proposes {\method}, a versatile and user-friendly \textbf{video-to-paragraph-to-video} editing method, supporting diverse video editing capabilities, such as removal, addition, and modification, through a unified pipeline. 
\method consists of two main stages: Video-to-Paragraph (V2P), which automatically generates structured descriptions of scene and object details, and Paragraph-to-Video (P2V), where users can refine these to guide a video diffusion model for flexible content edits, including removing, changing, or adding objects.
Key contributions of \method include: (1) A multi-granular spatiotemporal pooling strategy for structured video understanding, capturing both global context and fine-grained object details to enable precise text-based video editing without complex human annotations.
% that captures both global context and object details for precise text-based editing without complex annotations.
(2) A video generative model fine-tuned on a curated video-paragraph-mask dataset for improved editing and inpainting. (3) The ability to generate new objects by forecasting motion via auto-generated mask planning.
% (2) A video generation model fine-tuned on a curated video-paragraph-mask dataset for improved editing and inpainting. (3) The ability to generate new objects by forecasting motion via auto-generated mask planning.
In the end, users can easily edit complex videos with RACCooN's automatic explanations and guidance.
We demonstrate its versatile capabilities in video-to-paragraph generation (up to $9.4\%p \uparrow$ improvement in human evaluations), video content editing (relative $49.7\% \downarrow$ in FVD).
% , and can be integrated with SoTA video generation models for further enhancement.
\end{abstract}

%% file: sections/1_intro.tex
\section{Introduction}
Recent advances in video generative models~\citep{yan2021videogpt, hong2022cogvideo, esser2023structure, mei2023vidm, chai2023stablevideo, blattmann2023stable, wu2023tune}, including Sora~\citep{2024sora}, have demonstrated remarkable capabilities in creating high-quality videos. 
Simultaneously, video editing models~\citep{geyer2023tokenflow,qi2023fatezero,wang2024videocomposer,yu2023inpaint,wu2024towards,zhang2023avid} have gained significant attention for enabling users to modify content using \textit{user-written} instructions. However, building a versatile, user-friendly framework that facilitates easy video modification for personal use remains challenging. 
% Video editing models have gained attention for enabling users to modify content using \textit{user-written} instructions. However, building a versatile, user-friendly framework remains challenging. Key issues include:
Key challenges are as follows:

1) Training a unified model for multiple editing tasks (e.g., add, remove, change objects) is difficult~\cite{yu2025veggie}, and most methods focus on narrow tasks, such as background inpainting~\citep{yu2023inpaint,wu2024towards}, or attribute editing~\citep{geyer2023tokenflow,qi2023fatezero,jeong2023ground}.

% Training a unified model for multiple editing tasks (e.g., add, remove, change objects) is difficult, and most methods focus on narrow tasks like inpainting or attribute edits.
2) The need for well-structured prompts that accurately describe videos and support diverse editing tasks is critical, as prompt quality directly impacts model performance. However, generating such prompts is costly and time-consuming, with quality varying by annotator expertise. While Multimodal Large Language Models (MLLMs)~\citep{liu2023visual,yang2023vid2seq,yu2024crema} have been explored for automatic video description, they often miss key details in complex scenes, limiting seamless and user-friendly editing.
% the necessity for well-structured textual prompts that accurately describe videos and can be edited to support diverse video editing skills.
% The quality of prompts critically influences the models' capabilities and the quality of their outputs. Generating detailed prompts is time-consuming and costly, and the quality varies depending on the expertise of the annotators. 
% Although Multimodal Large Language Models (MLLMs)~\citep{liu2023visual,munasinghe2023pg,yang2023vid2seq,yu2024crema} have been explored for automatically describing videos, they often overlook crucial details in complex scenes, limiting seamless pipelines and user-friendly editing.

% High-quality, editable textual prompts are essential but costly and vary in quality. While MLLMs have been explored for automatic descriptions, they often miss crucial details in complex scenes, limiting seamless pipelines and user-friendly editing.

To tackle these limitations, we introduce {\method}: \textit{A Versatile Instructional Video Editing with Auto-Generated Narratives}, a novel \textbf{video-to-paragraph-to-video (V2P2V)} generative method that facilitates diverse video editing ({R}emove, {A}dd, and {C}hange) capabilities based on auto-generated video descriptions. \method enables seamless removal or modification of subject attributes and addition of new objects in videos \textbf{without the need for densely annotated prompts or extensive user input}. 
\method operates in two main stages: \textit{video-to-paragraph} (V2P) and \textit{paragraph-to-video} (P2V). 
In the V2P stage, we introduce a new video descriptive module built on a pre-trained Video-LLM backbone (PG-Video-LLaVA~\citep{munasinghe2023pg}). 
We find that existing Video-LLMs effectively capture holistic video features, yet often overlook detailed cues that are critical for accurate video editing, as users may be interested in altering these missing contexts. 
To address this, we propose a novel multi-granular video perception strategy that leverages superpixels~\citep{li2015superpixel,ke2023learning} to capture diverse and informative localized contexts throughout a video. We first extract fine-grained superpixels using a lightweight predictor~\citep{yang2020superpixel} and then apply overlapping k-means clustering~\citep{cleuziou2007generalization, whang2015non} to segment visual scenes into various levels of granularity. The suggested localized spatiotemporal segmentation assists the LLM's comprehension of objects, actions, and events within the video, enabling it to generate fluent and detailed natural language descriptions.
Next, in the P2V stage, we fine-tune a video inpainting model to support multiple editing tasks using auto-generated detailed descriptions and object masks.
Then, by leveraging user prompts derived from generated descriptions in the V2P stage, our video diffusion model accurately \textit{paint} corresponding video regions, ensuring that textual edits are faithfully reflected across various editing tasks. 

To further support model training, we introduce the \textbf{V}ideo \textbf{P}aragraph with \textbf{L}ocalized \textbf{M}ask (\textbf{VPLM}) dataset comprising over 7.2K quality video-paragraph data and 5.5K detailed object-level captions with masks, annotated from public datasets using GPT-4V~\citep{achiam2023gpt}. 

We emphasize that \method enhances the quality and versatility of video editing by leveraging detailed, automatically generated prompts that minimize ambiguity and refine the scope of generation.
We validate the extensive capabilities of the \method in both V2P generation, text-based video content editing, and video generation on ActivityNet~\citep{krishna2017dense}, YouCook2~\citep{zhou2018towards}, UCF101~\citep{soomro2012ucf101}, DAVIS~\citep{pont20172017}, and our proposed VPLM datasets. 
On the V2P side, \method outperforms several strong video captioning baselines~\citep{li2023videochat,munasinghe2023pg,liu2023visual}, particularly improving by average \textbf{+9.1\%p} on VPLM and up to \textbf{+9.4\%p} on YouCook2 compared to PG-VL~\citep{munasinghe2023pg}, based on both automatic metrics and human evaluation.
On the P2V side, \method surpasses previous strong video editing/inpainting baselines~\citep{geyer2023tokenflow,qi2023fatezero,wang2024videocomposer,yu2023inpaint,wu2024towards} over three subtasks of video content editing (remove, add, and change video objects) over 9 metrics. 
% We also demonstrate that the proposed \method can enhance SoTA video generative models by leveraging detailed auto-generated prompts. We further conduct extensive ablation and visualizations to validate the improvement quantitatively and qualitatively. 
Our contributions are as follows:
\vspace{-0.05in}
\begin{itemize}[leftmargin=1em]
\item[1.] {\textbf{Framework Contribution:} RACCooN offers a user-friendly, unified framework for diverse video editing tasks, enhancing interpretability and interaction by generating detailed, object-centric descriptions and layout plans tailored to editing goals, surpassing existing model combinations.}

\item[2.] {\textbf{Technical Contribution:} We present a novel \textbf{multi-granular pooling} strategy to capture local video contexts, enhancing video comprehension by generating fluent and detailed descriptions in a zero-shot setting. This enables users to create new videos that retain the visual characteristics of the input and support targeted context editing.}
\item[3.] {\textbf{Training/Dataset Contribution:} To enable RACCooN to handle complex and diverse video editing requests, we present the \textbf{VPLM} dataset, comprising 7.2K high-quality video paragraphs and 5.5K object-level caption-mask pairs. These well-structured annotations enable accurate V2P and P2V stages at both video and object levels.}
\end{itemize}

% we find high-quality & detailed textural prompt benefits for accurate video generation
% Our VPLM dataset makes an important initial step for enhancing video generation with high-quality, localized T2V data

% \begin{itemize}[leftmargin=2em]
% \item[1.] We introduce {\method}, a versatile and user-friendly \textbf{video-to-paragraph-to-video} generative framework that describes the input video with detailed natural language, which in turn enables users to create or edit videos without the need for any human-annotated input prompts.
% \item[2.] We present a \textbf{multi-granular pooling} strategy to capture local video contexts, enhancing video comprehension by generating fluent and detailed descriptions. This enables users to create new videos that retain the visual characteristics of the input and focus on specific context editing.
% \item[3.] We present the \textbf{VPLM} dataset, which contains 7.2K high-quality detailed video paragraphs, and 5.5K object-level detailed caption-mask pairs, facilitating the accurate V2P and P2V.

% \item[4.] We show the effectiveness and efficiency of \method framework via \textbf{extensive quantitative and qualitative analyses}, comparing with strong baselines in video captioning and editing.
% \end{itemize}

%% file: sections/2_related.tex
\input{materials/3_concept_figure}

\section{Related Work}

\textbf{Video-to-Paragraph Generation.}
Recent video-language tasks focus on generating comprehensive textual descriptions for long and complex video content~\citep{shen2017weakly,krishna2017dense,wang2018reconstruction,tewel2022zero,wu2023cap4video}. 
Vid2Seq~\citep{yang2023vid2seq} introduces a novel dense event captioning approach for narrated videos, with time tokens and event boundaries. 
Video-LLaVA variants~\citep{lin2023video,munasinghe2023pg} present a large multimodal model integrating text, video, and audio inputs for generative and question-answering tasks. 
Similarly, LLaVA-Next~\citep{zhang2024llavanextvideo} improves zero-shot video understanding by transferring multi-image knowledge through concatenated visual tokens.
While these methods are effective in video description, they often miss key contextual details~\citep{zhang2023video, li2023videochat}. \method captures both holistic and object-level details by leveraging localized spatiotemporal information, enhancing video editing and generation.

\textbf{Prompt-to-Video Editing.}
Video editing~\citep{ceylan2023pix2video, liu2023video, couairon2023videdit, kondratyuk2023videopoet, wang2023zero, zhang2024towards} involves enhancing, modifying, or manipulating video content for desired effects. VideoComposer~\citep{wang2024videocomposer} offers a multi-source controllable video generative framework. 
TokenFlow~\citep{geyer2023tokenflow} adapts text-to-image diffusion with flow matching for consistent text-driven video editing. 
LGVI~\citep{wu2024towards} integrates an MLLM for complex language-based video inpainting.
These methods often focus on specific tasks and may inadvertently alter unrelated regions due to limited contextual information. Our V2P2V framework overcomes these limitations by using auto-generated, detailed descriptions to integrate key contexts into diverse editing tasks.

%% file: materials/3_concept_figure.tex
\begin{figure*}
    \centering
    {\includegraphics[width=1\linewidth]{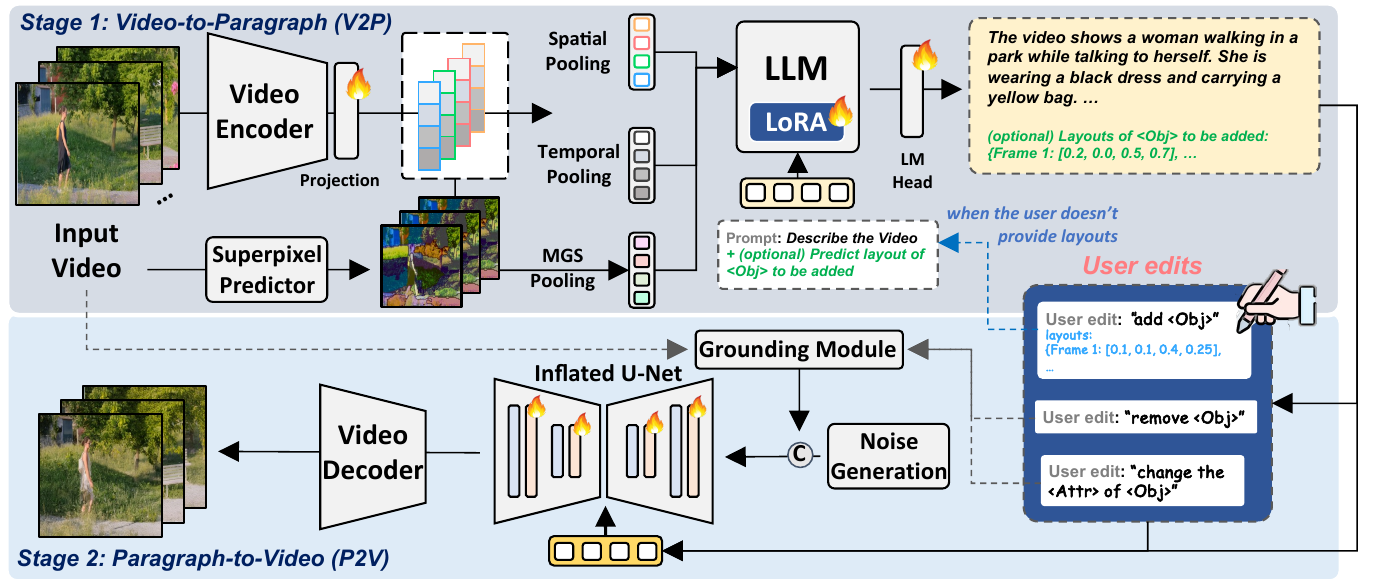}}
    \caption{
    \textbf{Illustration of {\method}.} 
    \method generates video descriptions with the three distinct pooled visual tokens, including Multi-Granular Spatiotemporal (MGS) Pooling. Next, users can edit the generated descriptions by adding, removing, or modifying words to create new videos. Note that for adding object tasks, if users do not provide layout information for the objects they want to add, \method can predict the target layout in each frame.}
    \label{fig:concept}
\end{figure*}

%% file: sections/3_method.tex
\section{A Versatile Instructional Video Editing with Auto-Generated Narratives}
\label{sec:method}
Conditional video generation and editing models struggle with complex scenes due to vague text descriptions and limited video understanding.
To address this, we introduce \method, a user-friendly, two-stage video-to-paragraph-to-video editing approach, with each stage detailed in \cref{sec:sub:v2p} and \cref{sec:sub:editing}.
% \method first generates structured paragraphs from videos using multi-granular spatiotemporal pooling to capture holistic content and key objects (\cref{sec:sub:v2p}).
% These auto-generated descriptions then guide conditional video generation and editing, enabling users to add, remove, or modify elements precisely and easily (\cref{sec:sub:editing}).
We also introduce the VPLM dataset, specifically curated to train models for detailed video editing, along with the training pipeline of \method, detailed in~\cref{sec:sub:traininig}.

\subsection{V2P: Auto-Descriptive Module with Multi-Granular Video Perception}\label{sec:sub:v2p}

\noindent\textbf{Multimodal LLM for Video Paragraph Generation.} 
In the V2P stage, the \method generates well-structured, detailed descriptions for both holistic videos and local objects. It employs a multimodal LLM with three main components: a visual encoder $E$, a multimodal projector, and an LLM. Given an input video $\bm{x}\in\mathbb{R}^{F\times C \times H\times W}$, where $F$, $C$, $H$, and $W$ represent the number of frames, channels, height, and width, respectively, we extract video features using the visual encoder: $\bm{e}=E(\bm{x})\in\mathbb{R}^{t\times h\times w \times d}$. Here, $t$, $h$, $w$, and $d$ denote the encoded temporal dimension, the height and width of the tokens, and the feature dimension.
To understand complex videos with multiple scenes, we use three pooling strategies: \textit{spatial pooling}, \textit{temporal pooling}, and \textit{multi-granular spatiotemporal pooling}. Spatial pooling $\bm{e}^s=\text{Pooling}^s(\bm{e})\in\mathbb{R}^{t\times d}$ aggregates tokens within the same frame, while temporal pooling $\bm{e}^t=\text{Pooling}^t(\bm{e})\in\mathbb{R}^{(h\cdot w) \times d}$ averages features across the temporal dimension for the same region. Despite these strategies helping the LLM grasp the video holistically in space or time, they often overlook capturing key objects or actions localized throughout the video stream, especially in untrimmed, dynamic, multi-scene videos.

\noindent\textbf{Multi-Granular Spatiotemporal Pooling.} 
To address this issue, we introduce a novel superpixel-based spatiotemporal pooling strategy, coined \textit{multi-granular spatiotemporal pooling} (MGS pooling). As illustrated in~\cref{fig:concept} left top, this strategy is designed to capture localized information via superpixels across spatial and temporal dimensions. Superpixels~\citep{li2015superpixel, yang2020superpixel, ke2023learning} are small and coherent clusters of pixels that share similar characteristics, such as color or texture. These clusters provide an efficient representation of visual scenes and are resilient to frame noise since they average out the pixel values within each cluster, effectively smoothing out variations induced by noise. 
As shown in~\cref{fig:mgspooling}, we use a lightweight superpixel predictor $\sigma(\cdot)$~\citep{yang2020superpixel} to generate superpixels across video frames, capturing the granular visuality of each local area. However, due to their limited coverage area, these fine-grained visual features often fail to capture attribute-level semantics, such as objects and actions~\citep{zhang2023video, li2023videochat}. 
Motivated by the importance of varying the compositions of multiple superpixels for different contexts in video understanding, we propose the use of overlapping k-means clustering~\citep{cleuziou2007generalization, whang2015non} for the obtained video superpixels, which improves the granularity from fine to coarse. This approach allows the LLM to gather informative cues about various objects and actions. We first obtain the pixel features and the superpixel index vector for the video pixels: $\bm{S}, \bm{g} = \sigma(\bm{x}, \bm{g}_{\text{init}})$, where $\bm{g}_{\text{init}}$ is the input superpixel indices, initialized by a region-based grid. 
Given the averaged pixel features of each superpixel, $\bar{\bm{S}}\in\mathbb{R}^{|\bm{g}|\times d_p}$, where $d_p$ denotes the pixel feature size, we generate the MGS tokens $\bm{e}^l$:
\begin{align}
\begin{split}
\bm{m}&=\textsc{OKM}\left(\bar{\bm{S}}, k, v\right)\in\{0,1\}^{k\times F\times H\times W},\\
\bm{e}^l&=\text{AvgPool}(\bm{m}) \otimes \bm{e}\in\mathbb{R}^{k\times d}, 
\label{eq:paragraph_generation}
\end{split}
\end{align}
where $\textsc{OKM}$ represents the overlapping k-means algorithm with $k$ centroids and overlap scale $v$ for each cluster. $\bm{m}$ denotes the set of binary masks for superpixels. $\otimes$ denotes tensor multiplication. We describe the detailed MGS process and ablation of pooling strategies in the Appendix.
Next, we concatenate the pooled video tokens and map them into the text embedding space using the multimodal linear projector. Combined with the embedding of the encoded text token $\bm{e}^{p}$ from the textual prompt, the LLM generates a well-structured and detailed description $\bm{a}$ of the video:
\begin{align}
\begin{split}
\widehat{\bm{e}}&=\textsc{concat}[\bm{e}^{s};\bm{e}^{l};\bm{e}^{t}] \cdot \bm{W}^\top, \\
\bm{a}&=LLM(\textsc{concat}[\bm{e}^{p}; \widehat{\bm{e}}]),
\label{eq:token_concat}
\end{split}
\end{align}
where $\bm{W}\in\mathbb{R}^{d \times d'}$ is the weight matrix for linear projection into the text embedding dimension $d'$. 
We highlight that our video description module serves as an integrated, user-interactive tool for video-to-paragraph generation and video content editing. (\cref{fig:concept} top right).

\input{materials/mgspooling}
\subsection{P2V: User-Interactive Video Editing with Auto-Generated Descriptions}\label{sec:sub:editing}
With the well-structured, detailed, and object-centric video description generated from the \textit{Video-to-Paragraph} stage, users can `read' the video details and interactively modify the content by altering the model-generated description. This approach shifts users' focus from labor-intensive video observation to content editing. We categorize general video content editing into three important subtasks:
\textbf{(1) Video Object Adding:} add extra objects to a video.
\textbf{(2) Video Object Removing:} delete target objects and re-generate the object region as the background.
\textbf{(3) Video Object Changing:} change objects' attributes (e.g., color, textural, material).
Many previous works have made great progress in video editing~\citep{wu2024towards, geyer2023tokenflow, qi2023fatezero, zhang2023avid,fan2024videoshop}, but usually focus on one of these subtasks. 
In this paper, we propose a unified generative model for video content editing that integrates all those crucial subtasks. Specifically, we formulate these subtasks as text-based video painting tasks and leverage a single video diffusion model for adding, removing, and changing video objects in the form of inpainting.

As shown in~\cref{fig:concept} bottom, our video diffusion model processes input video $\bm{x}\in\mathbb{R}^{F\times C \times H\times W}$ with a predicted binary mask $\bm{m}'\in\mathbb{R}^{F\times 1 \times H\times W}$ targeting specific regions for modification. Following image inpainting techniques~\citep{xie2023smartbrush, rombach2022high}, we apply the mask\footnote{We use image grounding~\citep{liu2023grounding} and video tracking models~\citep{cheng2023segment} as the off-the-shelf mask predictor in inference.} to the video to designate the editing region. The masked video is then encoded using a Variational Autoencoder (VAE~\citep{kingma2013auto}) to serve as the generation condition.
The model can then be informed on which video region should be edited for localized editing.  
Driven by the detailed description, the diffusion model can conduct diverse video editing that reflects the text prompts. 

{
In addition, we provide details regarding the process of adding objects in video editing. Indeed, adding objects can be considered a unique video editing task, distinct from removing objects or changing attributes. Unlike the latter scenarios, where the target objects are already present in the initial video, adding objects involves introducing entirely new elements, which necessitates a modified editing process.}
As illustrated in~\cref{fig:concept}, the MLLM in V2P provides not only detailed descriptions but also frame-wise placement suggestions for new objects in the form of bounding box sequences. The object insertion process in RACCooN in inference is conducted through the following steps:\\
\textbf{\textit{1. User Edit:}} The user provides an instruction to add a specific object.\\
\textbf{\textit{2. MLLM Output:}} The finetuned MLLM in V2P generates fine-grained video descriptions along with frame-wise bounding box suggestions for new objects. For example, “Layouts of <Obj> to be added: \{Frame 1: \texttt{[0.2, 0.0, 0.5, 0.7]}, Frame 2: \texttt{[0.2, 0.1, 0.4, 0.65]}, …\}” specifies the layout for each frame, where \texttt{[x1, y1, x2, y2]} represents the top-left and bottom-right corners of the bounding box, with coordinates normalized to the range $[0, 1]$ (yellow box in~\cref{fig:concept}, top right).\\
\textbf{\textit{3. Video Editing:}} Generate videos based on the MLLM-generated output, including the frame-wise layout of the object to be added.

\subsection{VPLM Dataset Collection and \method Pipeline Training}\label{sec:sub:traininig}

\noindent\textbf{Dataset Collection.} We utilize video datasets~\citep{majumdar2020improving,gavrilyuk2018actor} from previous video inpainting work~\citep{wu2024towards}. Each raw video is accompanied by multiple inpainted versions with specific objects removed and includes binary masks of these objects. 
Although well-annotated with object masks and inpainted backgrounds, these datasets lack detailed descriptions of holistic video and specific local objects, hindering \method's training for producing well-structured captions for video editing.
To address this, we use GPT-4V~\citep{achiam2023gpt} to annotate detailed video descriptions. 
We first re-arrange uniformly sampled video frames into a grid-image~\citep{fan2021image} and add visual prompts by numbering each frame. We then ask GPT-4V to generate detailed captions for both the entire video and key objects, in a well-structured format. 
Next, we train V2P and P2V stages in our pipeline separately (\cref{fig:concept}). 
In the end, \method can automatically generate detailed, well-structured descriptions for raw videos and adapt these descriptions based on user updates for various video content editing tasks.

\noindent\textbf{MLLM Instructional Fine-tuning.} 
To enable the MLLM to output detailed video descriptions for content editing, we construct an instructional fine-tuning dataset based on VPLM with two video-instruction~\citep{liu2023visual} designs: (1) For object editing and removal, the MLLM generates structured video captions identifying key objects in the original video $\bm{x}$, using annotated descriptions as the learning objective. 
This allows users to edit videos directly from these descriptions without exhaustive analysis.
(2) For object insertion, the MLLM provides not only detailed descriptions but also frame-wise placement suggestions for new objects, enhancing its utility in video editing by avoiding manual trajectory outlining.
For training, we convert video object segmentation masks into bounding boxes by selecting maximal and minimal coordinates and follow the box planning strategy using LLMs~\citep{lin2023videodirectorgpt}. We input box coordinates as a sequence of numbers and train \method to predict these layouts given inpainted videos $\widehat{\bm{x}}$. We perform parameter-efficient fine-tuning with LoRA~\citep{hu2021lora}\footnote{We employ LoRA for query and value for each self-attention.} on these mixed datasets with CE loss. We freeze the visual encoder and LLM backbone, updating the projector, LoRA, and LLM head.

\noindent\textbf{Video Diffusion Model Fine-tuning.} 
Our video diffusion model builds on the prior image inpainting model~\citep{rombach2022high}, enhanced with temporal attention layers to capture video dynamics. The model is designed to generate video that aligns with input prompts, focusing on object-centric video content editing.
To support this, we develop a training dataset of mask-object-description triples. We use GPT-4 to produce single-object descriptions from long, detailed video narratives, framing this task as a multi-choice QA problem. 
Next, for the three video editing subtasks, we design specific input-output combinations:
(1) \textbf{Video Object Addition:} Inputs: inpainted video $\widehat{\bm{x}}$, object bounding boxes from segmentation masks $m$, and detailed object description $\bm{p}$. Output: original video $\bm{x}$.
(2) \textbf{Video Object Removal:} Inputs: original video $\bm{x}$, object segmentation masks $m$, and a fixed background prompt. Output: inpainted video $\widehat{\bm{x}}$.
(3) \textbf{Video Object Change:} Inputs: original video $\bm{x}$, object segmentation masks $m$, and object description $\bm{p}$. Output: original video $\bm{x}$.
The model is fine-tuned following the prior work~\citep{wu2023tune}, updating only the temporal layers and the query projections within the self-attention and cross-attention modules. We employ the MSE loss between generated and random noise. See Appendix for more details on the dataset and training.

%% file: materials/mgspooling.tex
\begin{figure}[ t]
\centering
\includegraphics[width=\linewidth]{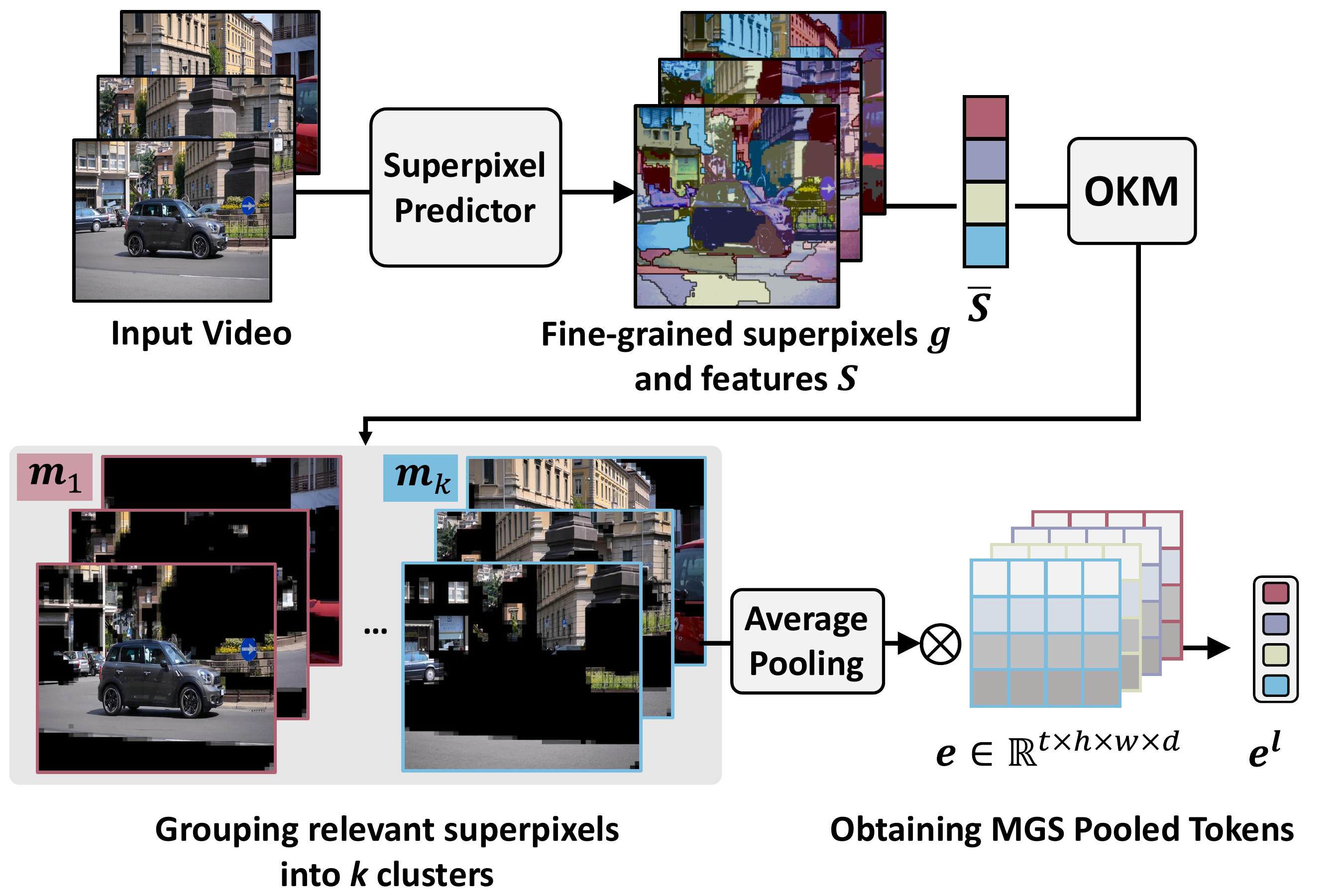}
\vspace{-0.25in}
\caption{
\textbf{Illustration of MGS pooling.} We obtain MGS pooling tokens using a spatiotemporal mask $\bm{m}$ via overlapping k-means clustering (OKM) of averaged superpixel features $\bar{\bm{S}}$.}
\vspace{-0.2in}
\label{fig:mgspooling}
\end{figure}

%% file: sections/4_exp.tex
\section{Experimental Results}\label{exp:results}
\input{materials/main_v_to_p}
\input{materials/video_content_editing}

% \subsection{Experiment Setup}\label{exp:setup}
\noindent\textbf{Tasks \& Datasets}: 
We evaluate \method on diverse video datasets across tasks, including video captioning (\textbf{YouCook2}~\citep{zhou2018towards}, \textbf{VPLM}), text-based video content editing (\textbf{DAVIS}~\citep{pont20172017}, \textbf{VPLM}), and conditional video generation (\textbf{ActivityNet}~\citep{krishna2017dense}, \textbf{YouCook2}~\citep{zhou2018towards}, \textbf{UCF101}~\citep{soomro2012ucf101}). 

\noindent\textbf{Metrics}: 
For each task, we evaluate our approach with various metrics. \textbf{(1) Video Caption:} following previous works~\citep{yang2023vid2seq, zhou2018end}, we conduct a comprehensive human evaluation and adopt general metrics for our long video descriptions, including SPICE~\citep{anderson2016spice}, BLEU-4~\citep{vedantam2015cider}, and CIDEr~\citep{vedantam2015cider}. 
\textbf{(2) Video Object Layout Planning:} following the prior work~\citep{lin2023videodirectorgpt}, we evaluate \method for object layout planning by bounding box IoU, FVD~\citep{unterthiner2019fvd}, and CLIP-score~\citep{radford2021learning}. 
\textbf{(3) Text-based Video Content Editing:} following prior works~\citep{geyer2023tokenflow,ceylan2023pix2video, yang2024eva}, we evaluate \method's video editing capabilities by CLIP-Text, CLIP-Frame, Qedit~\citep{yang2024eva}, and SSIM~\citep{hore2010image}. 
\textbf{(4) Conditional Video Generation:} we measure FVD~\citep{unterthiner2019fvd}, CLIP-Score~\citep{radford2021learning}, and SSIM~\citep{hore2010image}. Implementation details are provided in the Appendix.

\subsection{Video-to-Paragraph Generation}\label{exp:v2p}
\noindent\textbf{Video-Paragraph Alignment.}
We conducted a quantitative evaluation of our proposed \method's video-to-paragraph generation capabilities, comparing it against strong baselines with a focus on object-centric captioning and object layout planning. The results, summarized in \cref{tab:single_obj_and_bbox}, show that open-source video-LLMs (e.g., PG-VL, Video-Chat), which have smaller LLMs ($<13B$ parameters), struggle with object-centric captioning and usually fail to generate layout planning.
This is primarily due to their lack of instructional fine-tuning and insufficient video detail modeling without multi-granular pooling. 
In contrast, \method demonstrates superior performance in both object-centric captioning and complex object layout planning, benefiting from the instructional tuning on our VPLM dataset. 
Additionally, our method achieves competitive performance with proprietary MLLMs (e.g., Gemini 1.5 Pro, GPT-4o) in key object captioning and layout planning, demonstrating its superior instruction following and generation quality.
\begin{figure*}[t]
\centering
\includegraphics[width=1\linewidth]{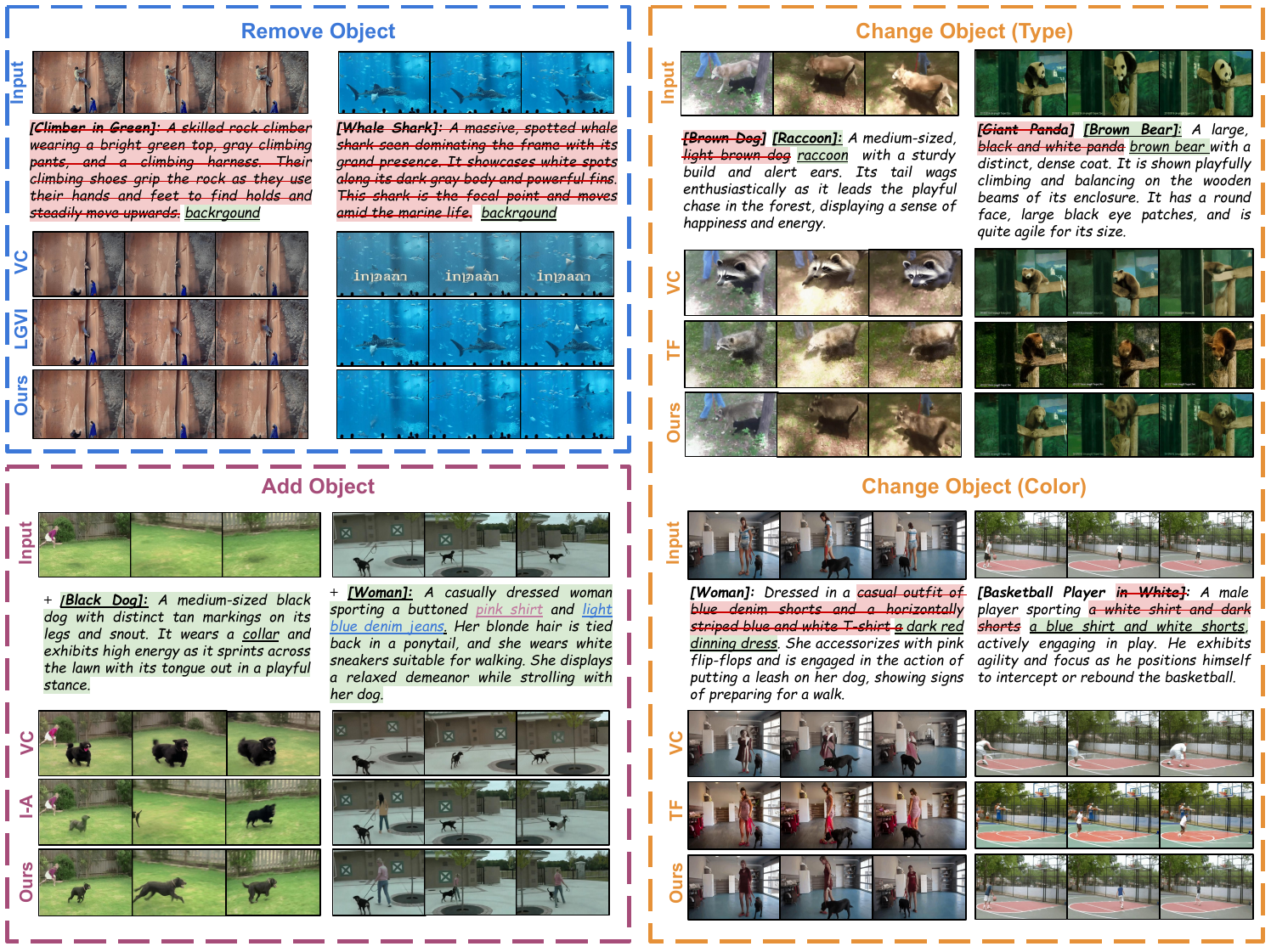}
\caption{\textbf{Qualitative Comparison between \method and other baselines.} Baseline names are abbreviated: \textbf{VC}: VideoComposer, \textbf{I-A}: Inpainting Anything, \textbf{TF}: TokenFlow. We \underline{underlined} visual details in our caption. More visualizations are in the supplementary material.} 
\label{fig:vis}
% \vspace{-0.1in}
\end{figure*}

\noindent\textbf{Human Evaluation \& Qualitative Examples.}
We conducted a human evaluation to compare our auto-generated captions with those from a strong baseline and human annotations on ten randomly selected YouCook2 videos (each three to five minutes long with multiple scenes and complex viewpoints). Five evaluators rated these based on Logic Fluency, Language Fluency, Video Summary, and Video Details (details in the supplementary material).
The average scores for each criterion and their overall mean are illustrated in \cref{tab:human_eval}. 
Our method significantly surpassed both the PG-VL-generated and ground truth captions in all metrics, showing a \textbf{4.9\%p} and \textbf{21.8\%p} \textbf{absolute improvement} respectively, and matched the ground truth in capturing \textit{Video Details} with a \textbf{9.4\%p} enhancement over the baseline, highlighting \method's superior capability in capturing video details.
We additionally visualize descriptions generated by \method using a well-known generated video from the Sora~\citep{2024sora} generated demo example. 
% As shown in \cref{fig:vis_para}, 
As shown in Fig. 4 (in the supplementary material), 
it demonstrates our robust capability to auto-describe complex video content without human textual input.

% \vspace{-1in}
\subsection{Instructional Video Editing with \method}\label{exp:p2v}
\noindent\textbf{Quantitative Evaluation.} 
As shown in~\cref{tab:video_content_editing}, we quantitatively compare the video editing ability of \method with strong video editing models based on inpainting or DDIM-inversion~\citep{hertz2022prompt} across three object-centric video content editing subtasks: \textit{object changing}, \textit{removal}, and \textit{adding}. In general, \method outperforms all baselines across 9 metrics.
For object changing, \method outperforms the best-performing baseline by 0.8\% on CLIP-T, indicating better video-text alignment while maintaining temporal consistency, as demonstrated by comparable CLIP-F and Qedit scores. Note that LGVI is not designed to alter video attributes and tends to preserve video content with marginal change (i.e., identical input and output videos), resulting in improved CLIP-F scores. In the object removal task, \method shows significant improvements over strong baselines (relatively $+57.8\%$ FVD, $+2.5\%$ SSIM, $+9.6\%$ PSNR). Such improvements are maintained in the addition task (relatively $+41.6\%$ FVD, $+4.3\%$ PSNR).
Meanwhile, some DDIM inversion-based models (e.g., TokenFlow~\citep{geyer2023tokenflow}) work well for specific tasks (change objects), but do not handle other types of editing. In contrast, our method is an all-rounder player. 
%that enables diverse video content editing skills. 

We further emphasize that \textbf{a simple combination of existing models cannot achieve an effective video editing pipeline}, leading to inferior instruction-following and editing abilities. This is evident in the degraded performance of open-source Video-LLM baselines (\cref{tab:single_obj_and_bbox,tab:human_eval}) and other video editing models (\cref{tab:video_content_editing}). To address these limitations, we made unique novelty in technical and dataset contributions and achieved significant improvements in both video understanding and editing. This is evident in the comparison of \method{} with a multi-agent baseline combining a powerful open-source video reasoning and video diffusion models (PG-Video-LLaMA + StableDiffusion 2.0-inpainting) in~\cref{tab:video_content_editing}. \method{} outperforms the \textit{PGVL + SD 2.0-inpainting} by a significant margin across all metrics and editing tasks, highlighting the effectiveness of our proposed editing method.

\noindent\textbf{{Visualization.}} 
In~\cref{fig:vis}, we compare videos generated by \method with several SoTA baselines across three video content editing tasks. For object removal, \method demonstrates superior results, naturally and smoothly inpainting the background, whereas VideoComposer generates unexpected content and LGVI fails to accurately remove objects across frames.
For object addition, compared to Inpainting-Anything and VideoComposer, which often miss objects or produce distorted generations, \method generates objects with more fluent and natural motion, accurately reflecting caption details (e.g., the \textit{collar} of the dog, the \textit{pink shirt}, and \textit{blue jeans} for the woman).
For changing objects, our method outperforms inpainting-based VideoComposer and inversion-based TokenFlow. \method accurately re-paints objects to achieve object editing for color (\textit{white}$\rightarrow$\textit{blue}) and type (\textit{dog}$\rightarrow$\textit{raccoon}), while others struggle to meet requirements.

\input{materials/video_content_editing_ablation}

\noindent\textbf{Ablation Studies.}
As shown in~\cref{tab:ablation_rm_add}, we further validate the effectiveness of components by replacing detailed descriptions with short captions, and oracle masks/planning boxes with model-generated ones. 
In adding objects, our detailed object descriptions can benefit generation by providing accurate details, leading to improved quantitative results (relatively $+14.4\%$ FVD). 
We further replace GT boxes with boxes predicted by \method, and still show superior performance over other baseline methods with oracle boxes in~\cref{tab:video_content_editing}. It demonstrates that our V2P stage can thus automatically generate planning from a given video to eliminate users' labor. 
Next, in object removal and changing, we replace the oracle masks with grounding~\citep{liu2023grounding} and tracking~\citep{cheng2023segment} tools generated mask, which shows a marginal decrement for changing objects, and \method still shows strong results over other baselines in~\cref{tab:video_content_editing} with oracle masks. 
It suggests that \method is effective and robust to handle diverse editing skills in a non-orcale setting (See the appendix).

%% file: materials/main_v_to_p.tex
\begin{table*}[t]
\begin{minipage}{.48\linewidth}
% \vspace{-0.1in}
\setlength{\tabcolsep}{2mm}
\resizebox{\textwidth}{!}{%
\begin{tabular}{lcccccc}
\toprule
\textbf{Methods} & \textbf{S} & \textbf{B} & \textbf{C} & \textbf{IoU} & \textbf{FVD} & \textbf{CLIP}\\ \midrule
\rowcolor{gray!20} 
\multicolumn{7}{c}{{\textit{open-source MLLMs}}}   \\
LLaVA~\citep{liu2023visual} & 17.4 &27.5&18.5&-&-&-\\ 
Video-Chat~\citep{li2023videochat} &  18.2&25.3&19.1&-&-&-\\ 
PG-VL~\citep{munasinghe2023pg} &18.2&27.4&14.6&-&-&-\\
\midrule
\rowcolor{gray!20} 
\multicolumn{7}{c}{{\textit{proprietary MLLMs}}}   \\
{Gemini 1.5 Pro~\citep{team2023gemini}}&{19.2}&{23.5}&{11.0}&{0.115}&{\textbf{371.63}}&{0.978}\\
{GPT-4o~\citep{gpt4o}}&
{20.6}&
{28.0}&
{\textbf{37.4}}&
{0.179}&
{447.67}&
{0.977}\\
\midrule
\textbf{\method} &  \textbf{23.1}&\textbf{31.0}&\underline{33.5}&
\textbf{0.218} &
\underline{432.42} &
\textbf{0.983} \\ 
\bottomrule
% \vspace{0.05in}
\end{tabular}
}
\vspace{-0.05in}
\caption{\textbf{Single Object Prediction} on VPLM test set. Metrics indicate:~\textbf{S}: \textit{SPICE}, \textbf{B}: \textit{BLEU-4}, \textbf{C}: \textit{CIDEr}.}\label{tab:single_obj_and_bbox}
\end{minipage}\hspace{0.15in}
\begin{minipage}{.48\linewidth}
  \centering
\resizebox{\textwidth}{!}{%
\begin{tabular}{lccccc}
\toprule
\textbf{Methods} & \textbf{Logic} & \textbf{Lang.} & \textbf{Summ.} & \textbf{Details} & \textbf{Avg.}\\
\midrule
Ground Truth &66.7&42.2&41.7&\textbf{72.2}&55.7\\
PG-VL~\citep{munasinghe2023pg} &77.2&81.1&69.4&62.8&72.6\\
\textbf{\method} &
\textbf{80.6}&
\textbf{85.0}&
\textbf{72.2}&
\textbf{72.2}&
\textbf{77.5}\\
\bottomrule
\vspace{0.05in}
\end{tabular}}\vspace{-0.1in}
\caption{\textbf{Results of Human Evaluation} on YouCook2. We measure the quality of the description through four metrics: Logic Fluency (Logic), Language Fluency (Lang.), Video Summary (Summ.), and Video Details (Details). We report the normalized score $s\in[0,100]$.}\label{tab:human_eval}
\vspace{0.1in}
\end{minipage} 
\end{table*}

%% file: materials/video_content_editing.tex
\begin{table*}[t]
\centering
\setlength{\tabcolsep}{1mm}
\resizebox{1\linewidth}{!}{%
\begin{tabular}{lccccccccccc}
\toprule
& \multicolumn{3}{c}{\textbf{Change Object}} &  & \multicolumn{3}{c}{\textbf{Remove Object}} &  & \multicolumn{3}{c}{\textbf{Add Object}} \\ \cmidrule(lr){2-4} \cmidrule(lr){6-8} \cmidrule(lr){10-12} 
\multirow{-2}{*}{\textbf{Model}} & \textbf{CLIP-T}$\uparrow$ & \textbf{CLIP-F}$\uparrow$ & \textbf{Qedit}$\uparrow$ &  & \textbf{FVD}$\downarrow$ & \textbf{SSIM}$\uparrow$ & \textbf{PSNR}$\uparrow$ &  & \textbf{FVD}$\downarrow$ & \textbf{SSIM}$\uparrow$ & \textbf{PSNR}$\uparrow$  \\ \midrule
\rowcolor{gray!20} \multicolumn{12}{c}{\textit{Inversion-based Models}} \\
{LOVECon}~\citep{liao2023lovecon} & {29.36} & {94.77}	& {1.29} &&	{1319.51}	&{60.40}&	{17.78}&&	{1433.12}	&{58.51}&	{17.35} \\
{FateZero}~\citep{qi2023fatezero}  & {25.18} &{94.47}&{1.01}&  &{1037.05}&{47.35}&{15.16}&     &{1474.80}&{47.65}&{15.45}\\
{TokenFlow}~\citep{geyer2023tokenflow} &{29.25}&{96.23}&{1.31}&  &{1317.29}& {47.06} &{15.83}& &{1373.20}&{49.95}&{15.95}\\
\rowcolor{gray!20} \multicolumn{12}{c}{\textit{Inpainting-Based Models}} \\
Inpaint Anything~\citep{yu2023inpaint} &24.86&92.01&1.01&  &383.81&82.33&27.69&   &712.59&77.75&22.41\\
LGVI~\citep{wu2024towards} &23.82&\textbf{95.33}&1.04&  &915.24&56.16& 19.14&  &1445.43&47.93&16.09\\ 
VideoComposer~\citep{wang2024videocomposer}  & 27.61 &94.18&\textbf{1.25}&   &827.04&47.34& 17.55&     &1151.90&48.01&15.76\\ 
% \midrule
PGVL + SD-v2.0-inpainting  & 24.01 & 90.11 & 1.01 & &282.31 & 82.33 & 27.69 & &1579.65 & 43.21 & 15.76\\ \midrule
\textbf{\method} &\textbf{27.85}& \underline{94.78}&\underline{1.15}&    &\textbf{162.03}&\textbf{84.38}&\textbf{30.34}&   &\textbf{415.82}&\textbf{77.81}&\textbf{23.38} \\
\bottomrule
\end{tabular}}
\caption{\textbf{Results of Video Content Editing on three sub-tasks} on VPLM test. We gray out models that conduct the DDIM inversion process and have a different focus on our inpainting-based model.}\label{tab:video_content_editing}
% \vspace{-0.1in}
\end{table*}

%% file: materials/video_content_editing_ablation.tex
% \begin{wraptable}{t}{0.7\linewidth}
\begin{table}[t]
% \vspace{-5mm}
\centering
\vspace{-0.1in}
\small
\setlength{\tabcolsep}{2mm}
\resizebox{1\linewidth}{!}{
\begin{tabular}{lccc}
\toprule
\textbf{Settings} & \multicolumn{1}{c}{\textbf{FVD}$\downarrow$} & \multicolumn{1}{c}{\textbf{SSIM}$\uparrow$} & \multicolumn{1}{c}{\textbf{PSNR}$\uparrow$} \\ \midrule
\rowcolor{gray!20} \multicolumn{4}{c}{\textit{add object}} \\
\method & 415.80 & 77.81 & 23.38 \\
w/o detail caption   & 476.01 & 76.80 & 23.14 \\
w/o oracle planning   & 969.95 & 76.65 & 21.21 \\ \midrule
\rowcolor{gray!20} \multicolumn{4}{c}{\textit{remove object}} \\
\method & 162.03 & 84.38 & 30.34 \\
w/o oracle mask & 398.01 & 81.60 & 27.15 \\ \midrule
\textbf{Setting} & \textbf{CLIP-T} & \textbf{CLIP-F}& \textbf{Qedit} \\ \midrule
\rowcolor{gray!20} \multicolumn{4}{c}{\textit{change object}} \\
\method & 27.85 & 94.78 & 1.15 \\
w/o oracle mask & 27.23 & 94.33 & 1.14 \\ \bottomrule
\end{tabular}}
\caption{\textbf{Ablation on video object changing, removing, and adding} with different inputs.}\label{tab:ablation_rm_add}
\vspace{-0.1in}
% \end{wraptable}
\end{table}

%% file: sections/5_conclusion.tex
\section{Conclusion}
We newly introduce an auto-descriptive video-to-paragraph-to-video generative approach. We automatically generate video descriptions by leveraging a multi-granular spatiotemporal pooling strategy, enhancing the model's ability to recognize detailed, localized video information. Our approach then uses these enriched descriptions to edit and generate video content, offering users the flexibility to modify content through textual updates, thus eliminating the need for detailed video annotations. Our video editing and generation abilities highlight notable effectiveness and enable a broader range of users to engage in video creation and editing tasks without the written textual prompts. 

\section*{Limitations}
\label{app:sec:limitation}
Our proposed \method has shown a remarkable ability to interpret input videos, producing well-structured and detailed descriptions that outperform strong video captioning baselines and even ground truths. However, it has the potential to produce inaccuracies or hallucination~\citep{liu2023mitigating, wang2024mitigating, zhou2024analyzing,ma2023world} in the generated text outputs.
In addition, the performance of \method in paragraph generation, video generation, and editing is influenced by the employed pre-trained backbones, including an LLM~\citep{touvron2023llama}, base Inpainting Model~\citep{rombach2022high}, Video Diffusion Model~\citep{xing2023dynamicrafter}, and Video Editor~\citep{geyer2023tokenflow}. However, our key contributions are independent of these backbones, and we emphasize that \method's capabilities can be further enhanced with future advancements in these generative model backbones. 

LLM-empowered video description and photorealistic video creation/editing inherit biases from their training data, leading to several broader impacts, including societal stereotypes, biased interpretation of actions, and privacy concerns. To mitigate these broader impacts, it is essential to carefully develop and implement generative and video description models, such as considering diversifying training datasets, implementing fairness and bias evaluation metrics, and engaging communities to understand and address their concerns. 

% The performance of our proposed framework in paragraph generation, video generation, and editing is influenced by the employed pre-trained backbones, including an LLM~\citep{touvron2023llama}.
% LLM-empowered video description and photorealistic video creation/editing inherit biases from their training data, leading to several potentially negative impacts, including societal stereotypes, biased interpretation of actions, and privacy concerns. To mitigate these potential negative impacts, it is essential to carefully develop and implement generative and video description models, such as considering diversifying training datasets, implementing fairness and bias evaluation metrics, and engaging communities to understand and address their concerns. 

% \section*{Reproducibility Statement}
% % To maximize reproducibility, we have included our code in the supplementary material. 
% % Also, we report all of our hyperparameter settings and model details in the Appendix. 
% This paper fully discloses all the information needed to reproduce the main experimental results
% of the paper to the extent that it affects the main claims and/or conclusions. To maximize reproducibility, we have included our code in the supplementary material. Also, we report all of our hyperparameter settings and model details in the Appendix.

\section*{Acknowledgement} 
We thank the reviewers and Jaemin Cho, Han Lin, Jialu Li, and Elias Stengel-Eskin for the thoughtful discussion. This work was supported by DARPA ECOLE Program No. HR00112390060, NSF-AI Engage Institute DRL-2112635, DARPA Machine Commonsense (MCS) Grant N66001-19-2-4031, ARO Award W911NF2110220, ONR Grant N00014-23-1-2356, and Accelerate Foundation Models Research program. The views contained in this article are those of the authors and not of the funding agency.

%% file: sections/6_appendix.tex
In this appendix, we present the following:
\begin{itemize}
    \item More details about VPLM dataset collection (\cref{app:vplm_collect}), experimental setups (\cref{app:sec:sub:datasets}), more implementation details (\cref{app:sec:sub:implementation}).
    % \item Limitations and Negative Societal Impact of \method (\cref{app:sec:limitation}).
    \item Additional analysis including ablations (\cref{app:sec:sub:mask_init}, \cref{app:sec:sub:ablation}).
    \item Additional qualitative examples with \method on video content editing (\cref{app:sec:sub:visualization}).
\end{itemize}

\section{Experimental Setup}\label{app:sec:setup}
\begin{figure*}
\centering
\includegraphics[width=0.9\linewidth]{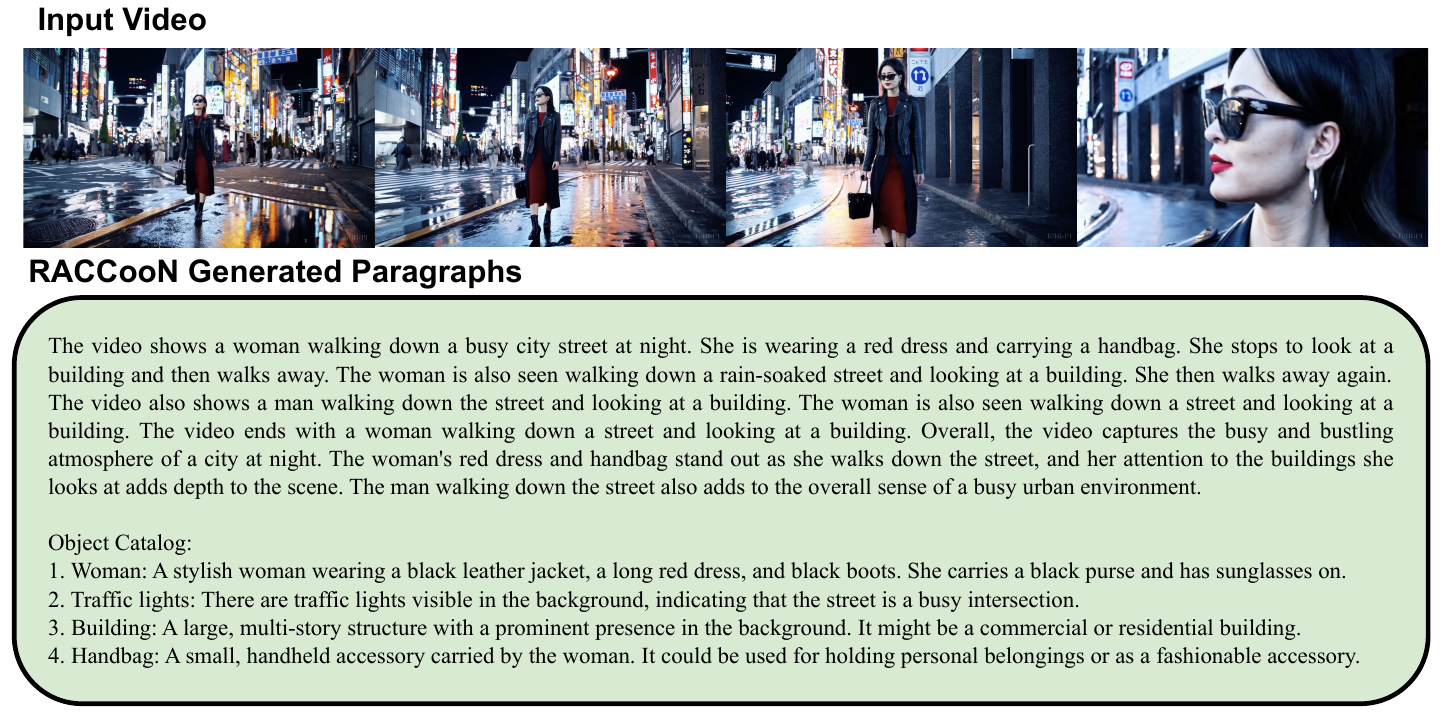}
\vspace{-0.05in}
\caption{\textbf{Qualitative V2P example of our \method} on Sora video.} 
\label{fig:vis_para}
% \vspace{-0.1in}
\end{figure*}
\vspace{0.1in}

\begin{figure*}[ht]
\centering
\vspace{-0.1in}
\includegraphics[width=\textwidth]{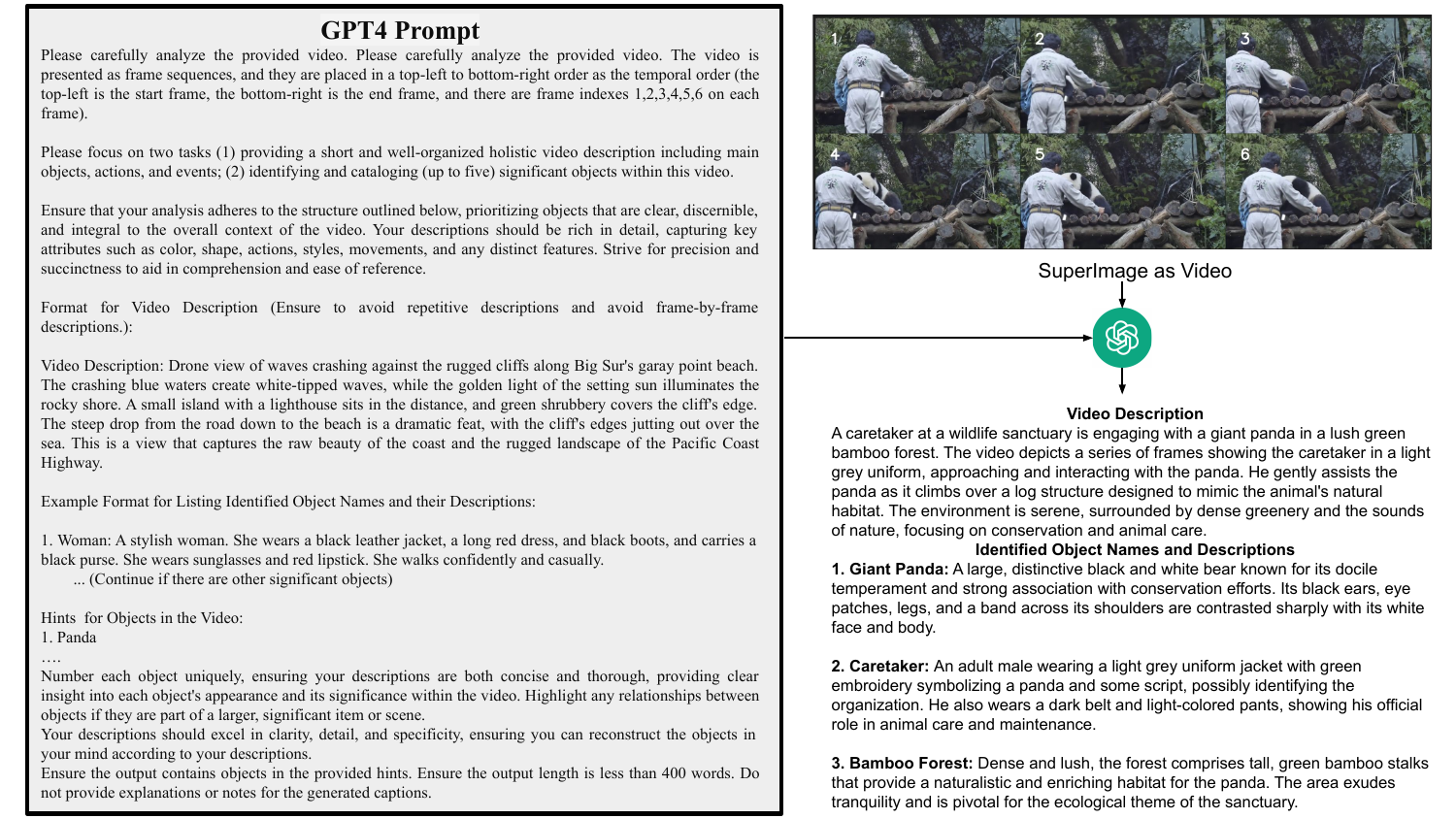}
\caption{
\textbf{Pipeline of our VPLM dataset annotation with GPT4V}. We first convert a video as a superimage and then give some in-context examples to prompt GPT-4V to annotate detailed and well-structured video descriptions.
}
\label{fig:annotation}
\vspace{-0.1in}
\end{figure*}

\subsection{VPLM data collection.}\label{app:vplm_collect}
As we mention in~\cref{sec:sub:traininig}, to facilitate our model training, we start from open-source video inpainting data~\citep{wu2024towards}\footnote{MIT License: \url{https://choosealicense.com/licenses/mit/}} to build a high-quality dataset that includes the well-structured, detailed caption for both video and each object in the video. 
Specifically, we leverage GPT-4V~\footnote{version 1106} to annotate each video. As shown in~\cref{fig:annotation}, we first convert a video to a superimage~\citep{fan2021image} by concatenating uniformly sampled frames, and we also draw frame IDs on each frame as a visual prompt to present temporal order. Then we prompt the GPT-4V by providing a detailed prompt with in-context examples (left of~\cref{fig:annotation}). In this case, we obtained well-structured, detailed captions that contain both holistic video and local objects (bottom right of~\cref{fig:annotation}). 
To ensure the annotation quality, we sampled 1 annotated video from each 100 batches and then did a human cross-check, and refined the batch annotations according to the sampled example. Through this pipeline, we obtained 7.2K high-quality quality well-structured, detailed video descriptions with an average of 238.0 words for each video.

Next, to obtain paired object-mask-description triplets for video inpainting model training, we build an automatic detailed object caption and object name matching pipeline using GPT4.
As in our base dataset~\citep{wu2024towards}, we already have class labels for each object mask,
we framed this matching as a multi-choice QA to ask GPT4 which object caption can in~\cref{fig:annotation} matched to the given object classes. We further filtered out the triplets with too small masks ($<1\%$ mask areas.)
In this case, we obtained 5.5K object-mask-description triplets with an average of 37.2 words for each object to support our video diffusion model training.

\subsection{Benchmarks and Datasets Details}\label{app:sec:sub:datasets}

As mentioned in the main paper~(\cref{exp:results}), we evaluate our proposed \method on various tasks.
For video-to-paragraph generation, we test our model on the standard video caption dataset YouCook2~\citep{zhou2018towards} (validation set) as well as our VPLM dataset. We next test video content editing with three subtasks on our VPLM dataset.
Regarding the experiments of incorporating \method with other conditional video generation models, we test \method on diverse videos from ActivityNet, YouCook2, and UCF101. 
We uniformly selected 100 videos from those 3 datasets to build the test bed. 
For the experiments of incorporating \method with other video editing models, we follow the previous work~\citep{geyer2023tokenflow}, and select 30 unique videos from the DAVIS dataset. For each video, we annotate two different types of editing, attribute editing and instance editing. 
It leads to 60 text-video pairs in our video editing evaluation. We choose object captions that contain the same keywords for editing in human captions to represent the model-generated caption.

\subsection{Implementation Details}\label{app:sec:sub:implementation}
% \noindent\textbf{Basics}: 
In V2P generation, we set $k=[20, 25]$ and $v=[5, 6]$ for superpixel clustering. We use CLIP-L/14@336~\citep{radford2021learning} as the image encoder and Vicuna-1.5~\citep{zheng2024judging} as the LLM. Our P2V model starts from StableDiffusion-2.0-Inpainting~\citep{rombach2022high}.
We split the VPLM datasets into train and test sets, with the test set containing 50 unique video-paragraph pairs (for V2P) and 180 mask-object-description triples (for P2V). We manually annotate the editing prompts for the object-changing subtask.
We quantitatively compare \method and other baselines on the VLPM test set. To focus on generation results rather than grounding ability, we apply the same ground truth masks and captions to all methods for P2V evaluation. 
See the supplementary material for more details on datasets, metrics, implementations, ablations, and qualitative analysis.

\noindent\textbf{Metrics}:
We provide more details about our metrics.
CLIP-Text measures the similarity between the edit prompt and the embedding of each frame in the edited video. 
CLIP-Frame computes the average CLIP similarity between the image embeddings of consecutive frames to measure the temporal coherence. 
SSIM measures the structural similarity between the original and edited video for evaluating localized editing.
$Q_{edit} = CLIP-T/Wrap-Err$, it is a comprehensive score for video editing quality, where $Warp-Err$ calculates the pixel-level difference by warping the edited video frames according to the estimated optical flow of the source video, extracted by FlowNet2.0~\citep{ilg2017flownet}.
For layout planning, we compute the FVD and CLIP-Image similarity between the ground truth and the predicted bounding box.

\noindent\textbf{More Details about Multi-granular Spatiotemporal Pooling.} As mentioned~\cref{sec:sub:v2p}, we proposed a novel Multi-granular Spatiotemporal Pooling (MGS Pooling) to address the lack of complex spatial-temporal modeling in video. We further provide a more intuitive visualization for our proposed MGS Pooling in~\cref{fig:mgspooling}. We first adopt a lightweight
165 superpixel predictor that generates superpixels across video frames, then use the overlapping k-means clustering for the obtained video superpixels. In this case, we gather informative cues about various objects and actions for LLM. 

\noindent\textbf{Human Evaluation on Video-to-Paragraph Generation.} We conduct a human evaluation on nine randomly selected videos from the YouCook2 dataset. 
Videos are three- to five-minute-long and contain multiple successive scenes with complex viewpoints. We provide these videos to four different annotators with ground truth captions, descriptions generated by PG-Video-LLAVA, and our method, \method, where the captions/descriptions for each video are randomly shuffled. We leverage four distinct human evaluation metrics: Logic Fluency (Logic), Language Fluency (Language), Video Summary (Summary) and Video Details (Details). To avoid a misinterpretation of methods' capabilities due to relative evaluation, we instruct the annotators to independently rate the quality of each set of captions based on these four different criteria, by giving a score from 1 to 5 (i.e., choice: [1, 2, 3, 4, 5]).

\noindent\textbf{Off-shelf Video Editing Models.} 
We utilize TokenFlow~\citep{geyer2023tokenflow} and FateZero~\citep{qi2023fatezero} as our video editing tools. 
TokenFlow generates a high-quality video corresponding to the target text, while preserving the spatial layout and motion of the input video. For SSIM computation, we compute SSIM for the region-of-no-interest since we want to keep those regions unchanged. 
We first mask out regions of interest with the ground truth mask provided by the DAVIS dataset, then we compute SSIM on masked images and conduct mean pooling as the video-level metrics. 

\noindent\textbf{Off-shelf Conditional Video Generation Models.} 
We leverage both VideoCrafter~\citep{chen2023videocrafter1} and DynamiCrafter~\citep{xing2023dynamicrafter} as our video generation backbone. 
VideoCrafter is one of the SoTA video generation models that can handle different input conditions (image, text).
DynamiCrafter is based on the open-source VideoCrafter and T2I Latent Diffusion model~\citep{rombach2022high}, and was trained on WebVid10M~\citep{bain2021frozen}, it provides better dynamics and stronger coherence.
We adopt VideoCrafter-512 and DynamiCrafter-512 variants. 
For each video, we use CLIP similarity to retrieve multiple keyframes corresponding to each caption. Those keyframes result in multiple generated video clips via the video generation model. 
For FVD computation, we conduct mean pooling over those clips to represent a video. 
We use $k=25$ and $v=6$ for generated captions in all experiments. The experiments are conducted on the 4 × 48GB A6000 GPUs machine. 

% \section{Limitations and Broader Impact}\label{app:sec:limitation}
% Our proposed \method has shown a remarkable ability to interpret input videos, producing well-structured and detailed descriptions that outperform strong video captioning baselines and even ground truths. However, it has the potential to produce inaccuracies or hallucination~\citep{liu2023mitigating, wang2024mitigating, zhou2024analyzing,ma2023world} in the generated text outputs.
% In addition, the performance of \method in paragraph generation, video generation, and editing is influenced by the employed pre-trained backbones, including an LLM~\citep{touvron2023llama}, base Inpainting Model~\citep{rombach2022high}, Video Diffusion Model~\citep{xing2023dynamicrafter}, and Video Editor~\citep{geyer2023tokenflow}. However, our key contributions are independent of these backbones, and we emphasize that \method's capabilities can be further enhanced with future advancements in these generative model backbones. 

% LLM-empowered video description and photorealistic video creation/editing inherit biases from their training data, leading to several broader impacts, including societal stereotypes, biased interpretation of actions, and privacy concerns. To mitigate these broader impacts, it is essential to carefully develop and implement generative and video description models, such as considering diversifying training datasets, implementing fairness and bias evaluation metrics, and engaging communities to understand and address their concerns. 

\section{Additional Analysis}\label{app:sec:analysis}

\subsection{Ablation study}\label{app:sec:sub:ablation}

\noindent{\textbf{The effect of $k$ and $v$}}
As shown in~\cref{tab:k_means_ablation}, we did initialized hyperparameter probing experiments on ActivityNet-Cap and YouCook2 datasets.
we observe that all variants of our approach with varying $k$ and $v$ generally achieve improved performance compared to baselines in terms of multiple video captioning metrics: \textit{SPICE}, \textit{BLEU-4}, \textit{METEOR}, and \textit{ROUGE}. This result demonstrates the efficacy of our multi-granular spatiotemporal pooling approach with a fine-to-coarse search of video contexts based on superpixels. In addition, we observe that \method shows a small gap between variants in each dataset, highlighting the robustness of our approach to the hyperparameter setups and datasets.

\input{materials/apdx_k_means_ablations}

\noindent{\textbf{The effect of Superpixel Overlap}} We introduce overlapping k-means clustering to aggregate video superpixels, capturing a variety of visual contexts while allowing for partial spatiotemporal overlap. To investigate the effect of our suggested overlapping approach, we also evaluate the variant of our approach without overlap (i.e., $v=1$) on video-to-paragraph generation tasks in~\cref{tab:k_means_ablation}. As shown, our approach with overlap (i.e., $v>1$) surpasses the non-overlapping version of \method across various scales of visual contexts $k$, as indicated by the video captioning metrics we evaluated. This emphasizes the advantage of permitting overlap in understanding video contexts, which enhances the input video's comprehension by allowing for diverse and fluent interpretations of local visual regions with surroundings associated at the same time.

For simplicity, we use $k=25$ and $v=6$ for all experiments on conditional video generation and video editing tasks, demonstrating the robustness of \method for hyperparameters.

\subsection{Comparison with Pre-trained Grounding Models}\label{app:sec:sub:mask_init} 

\input{materials/apdx_spixel_init}

We further investigate the applicability of recent powerful pre-trained visual grounding models~\citep{kirillov2023segany, cheng2023segment, ren2024grounded}. Segment-Anything~\citep{kirillov2023segany} and Grounded SAM~\citep{ren2024grounded} are strong open-ended object segmentation models for images, and we directly compute our localized granular tokens based on their segmentation masks. We select $25$ segmentation masks in total, from uniformly sampled frames in each video for fair comparison with \method ($k=25$). As shown in \cref{tab:grounding}, these variants of \method achieve improved performance against the best-performing baseline, PG-VL, but are often suboptimal since they focus on regional information and cannot contain the temporal information of the videos. Unlike these image-based segmentation methods, SAM-Track~\citep{cheng2023segment} generates coherent masks of observed objects over successive frames in videos, by adopting multiple additional pre-trained modules, including GroundingDino~\citep{liu2023grounding} and AOT~\citep{yang2021aot,yang2022deaot}. 
We adopt SAM-Track to initialize superpixels in videos and conduct overlapping k-means clustering ($k=25$). Here, we observe that \method with SAM-Track superpixel initialization achieves reasonable performance, and is beneficial for video editing tasks. 
It enables the model to coherently edit targeted regions in videos with edited keywords.

\subsection{Additional Visualizations}\label{app:sec:sub:visualization}
In this section, we provide more qualitative examples of various tasks, including three types of video content editing, ablation on removing oracle planning and GT masks, enhanced video editing, and conditional video generation.

\begin{figure*}
    \centering
    {\includegraphics[width=0.9\textwidth]{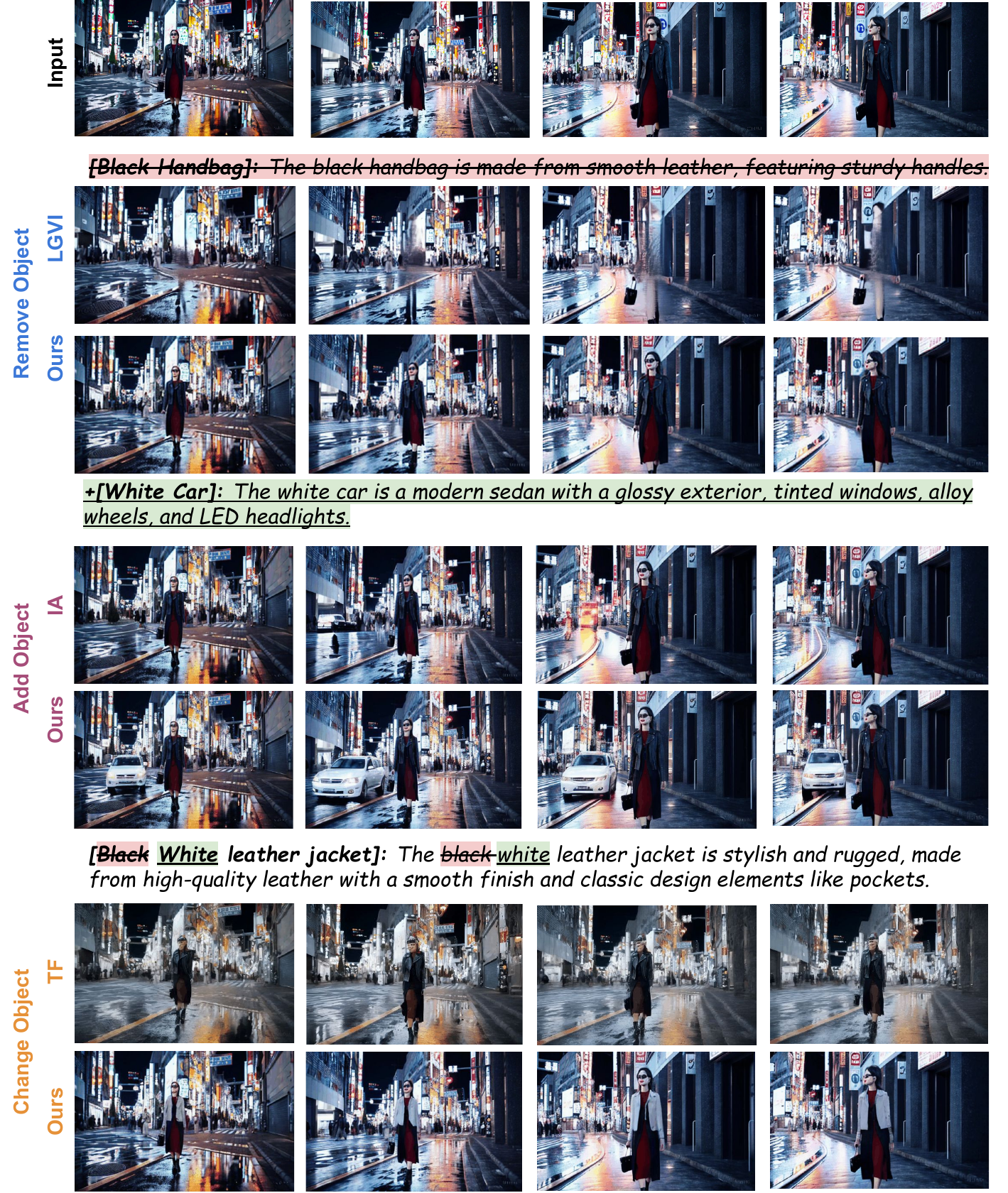}}
    \caption{More visualization of diverse editing skills on Sora video and comparison with other methods}
    \label{fig:sora_example}
\end{figure*}

\begin{figure*}
    \centering
    {\includegraphics[width=0.9\textwidth]{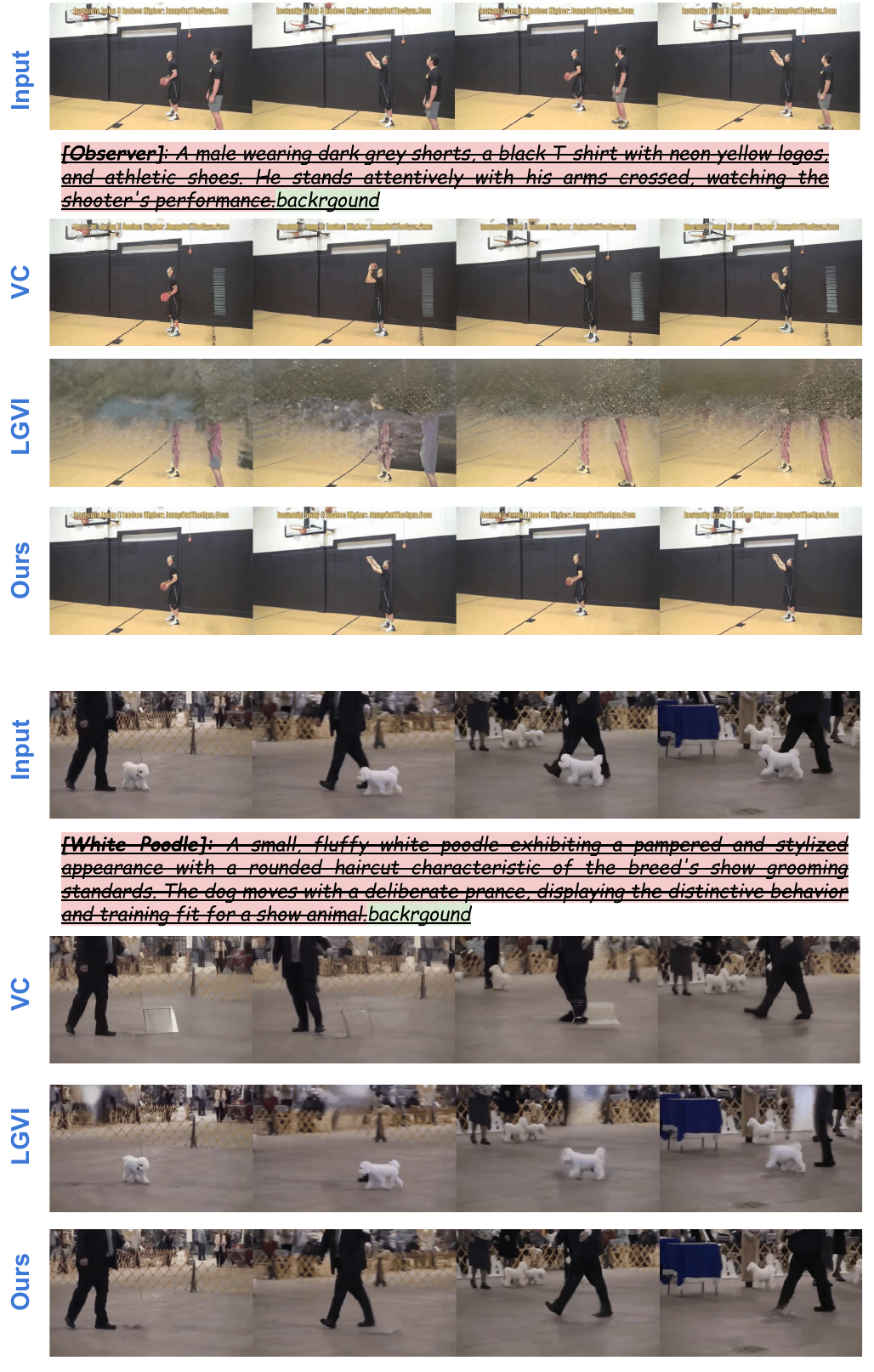}}
    \caption{More visualization of \textbf{removing} video objects}
    \label{fig:remove1}
\end{figure*}

\begin{figure*}
    \centering
    {\includegraphics[width=0.9\textwidth]{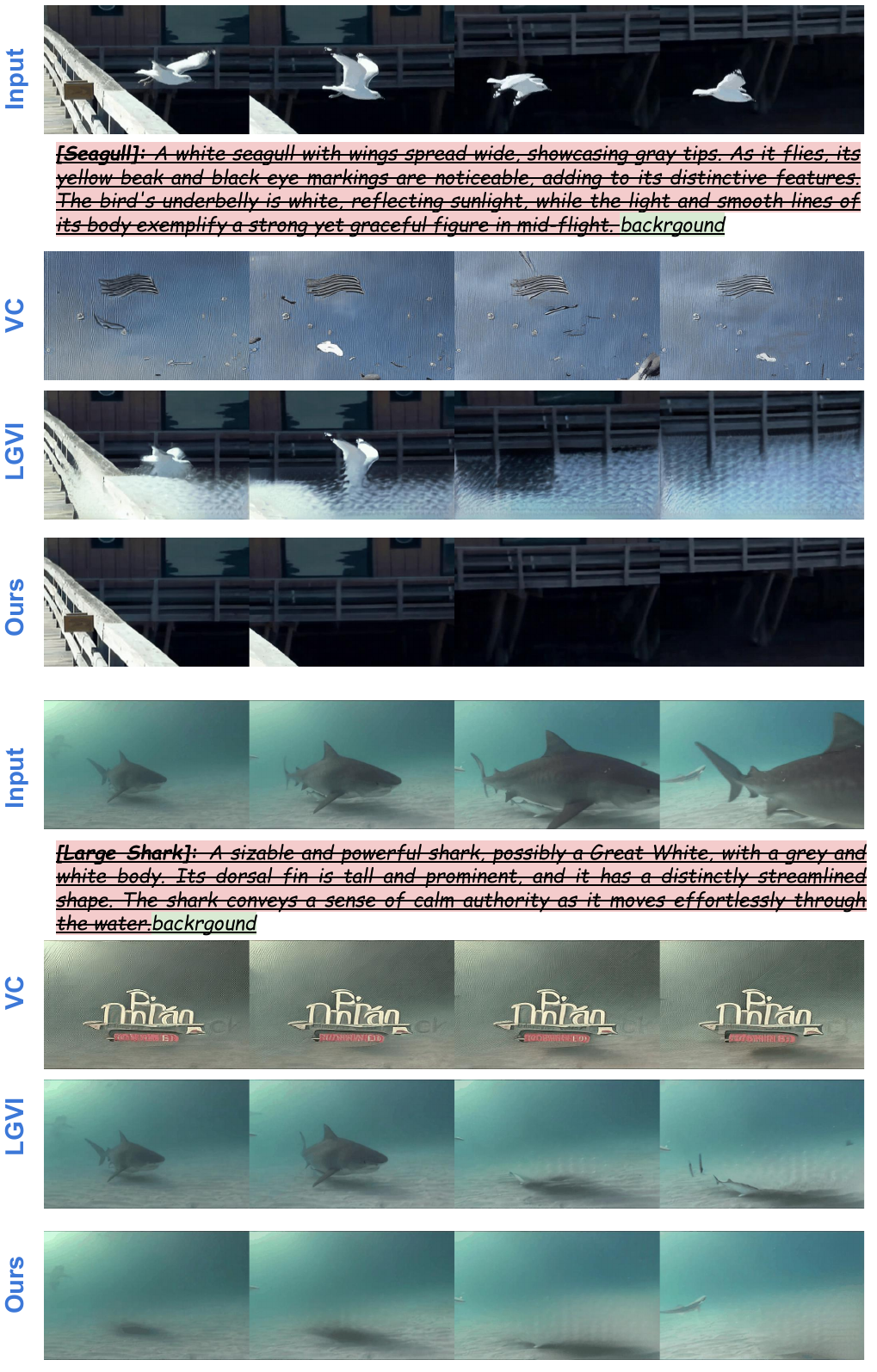}}
    \caption{More visualization of \textbf{removing} video objects}
    \label{fig:remove2}
\end{figure*}

\begin{figure*}
    \centering
    {\includegraphics[width=0.9\textwidth]{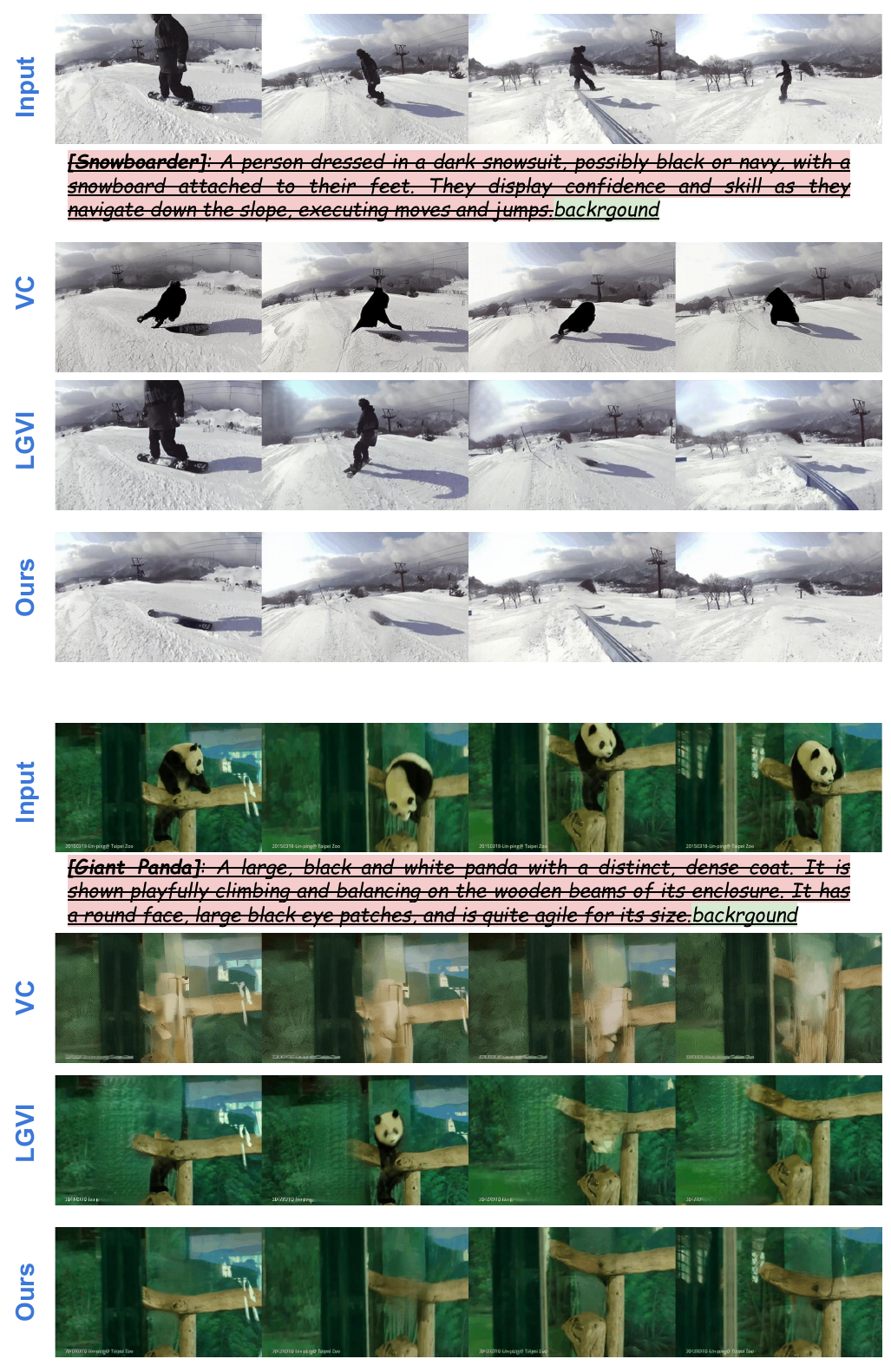}}
    \caption{More visualization of \textbf{removing} video objects}
    \label{fig:remove3}
\end{figure*}

\begin{figure*}
    \centering
    {\includegraphics[width=0.9\textwidth]{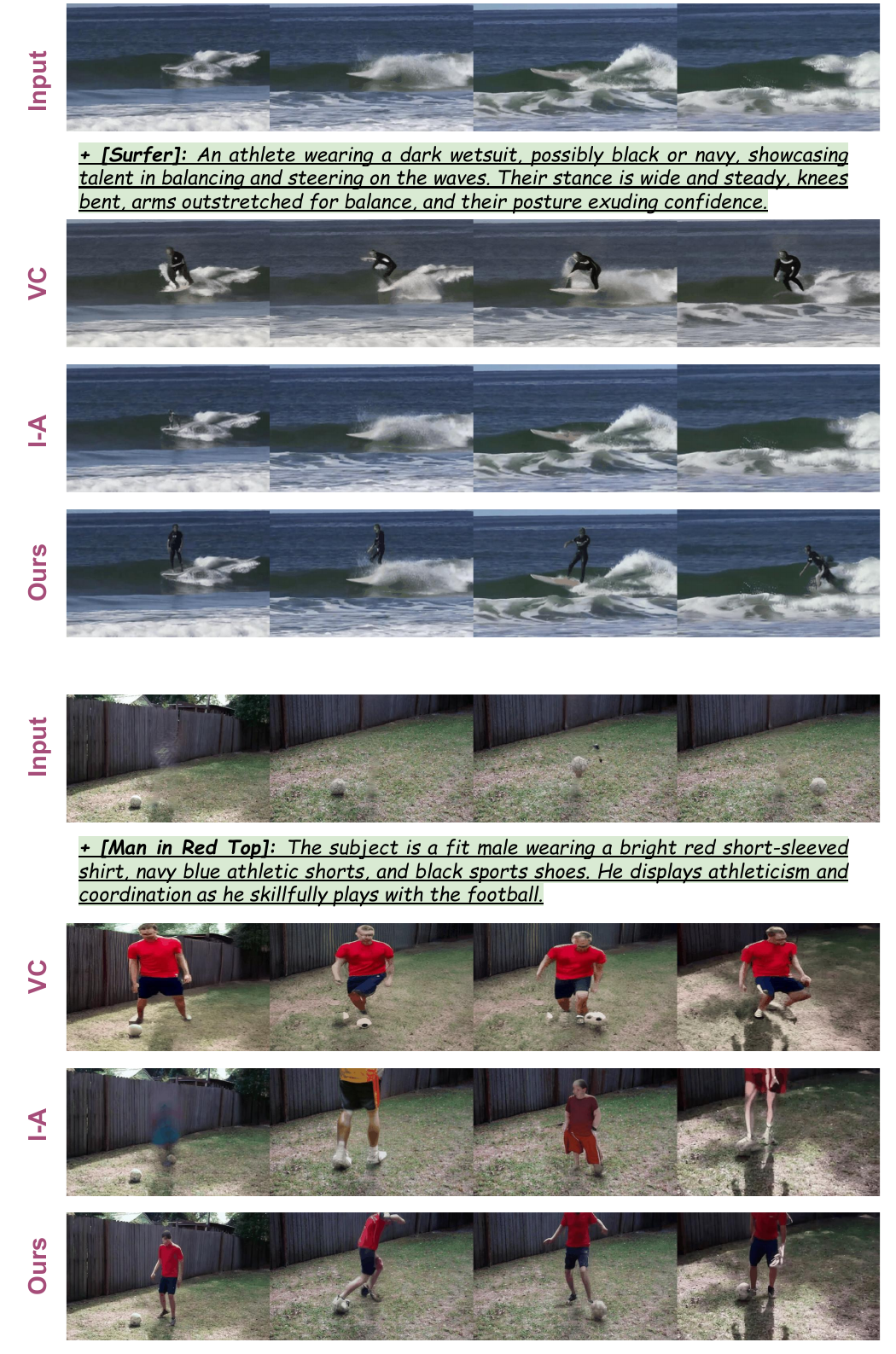}}
    \caption{More visualization of \textbf{adding} video objects}
    \label{fig:add1}
\end{figure*}

\begin{figure*}
    \centering
    {\includegraphics[width=0.9\textwidth]{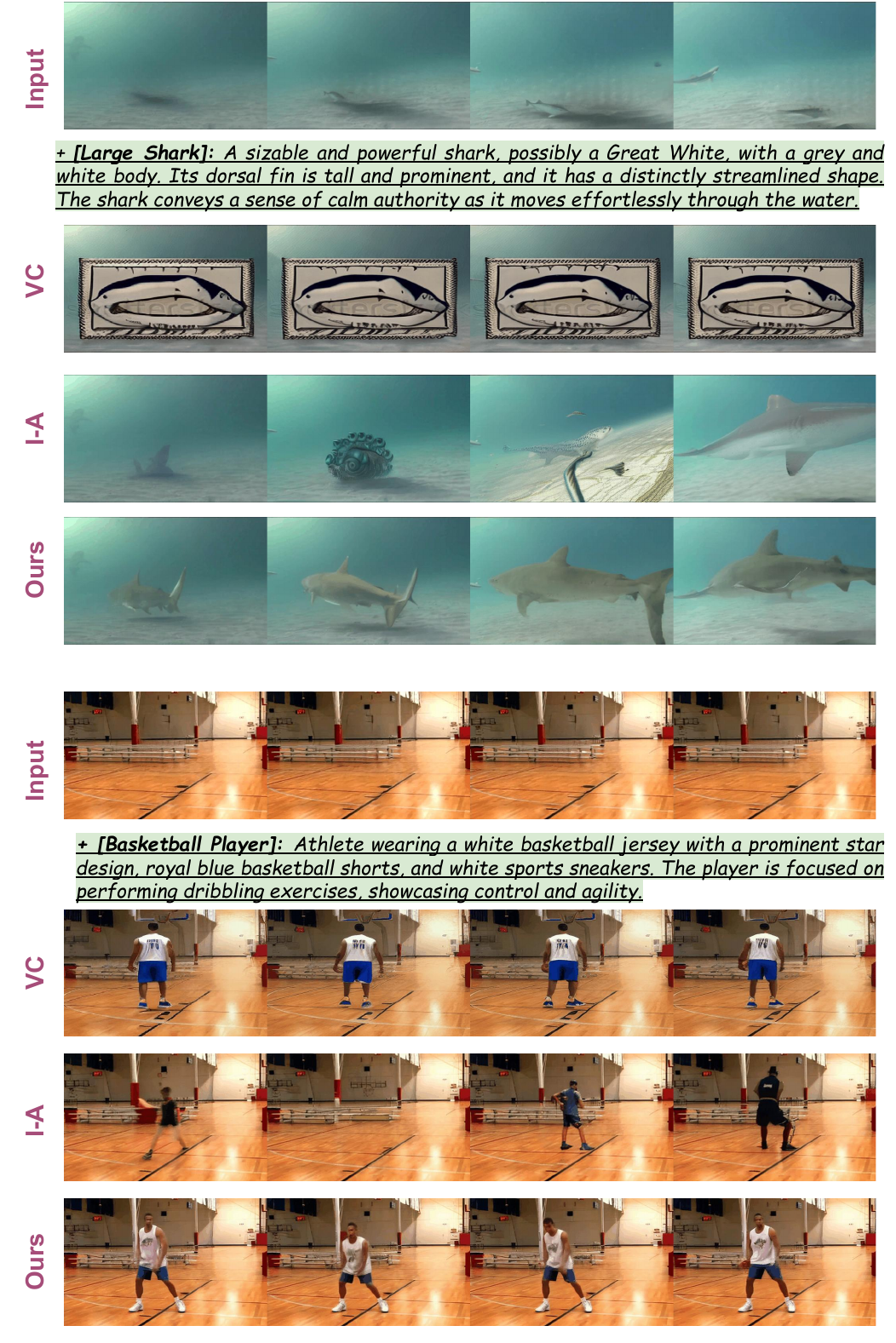}}
    \caption{More visualization of \textbf{adding} video objects}
    \label{fig:add2}
\end{figure*}

\begin{figure*}
    \centering
    {\includegraphics[width=0.9\textwidth]{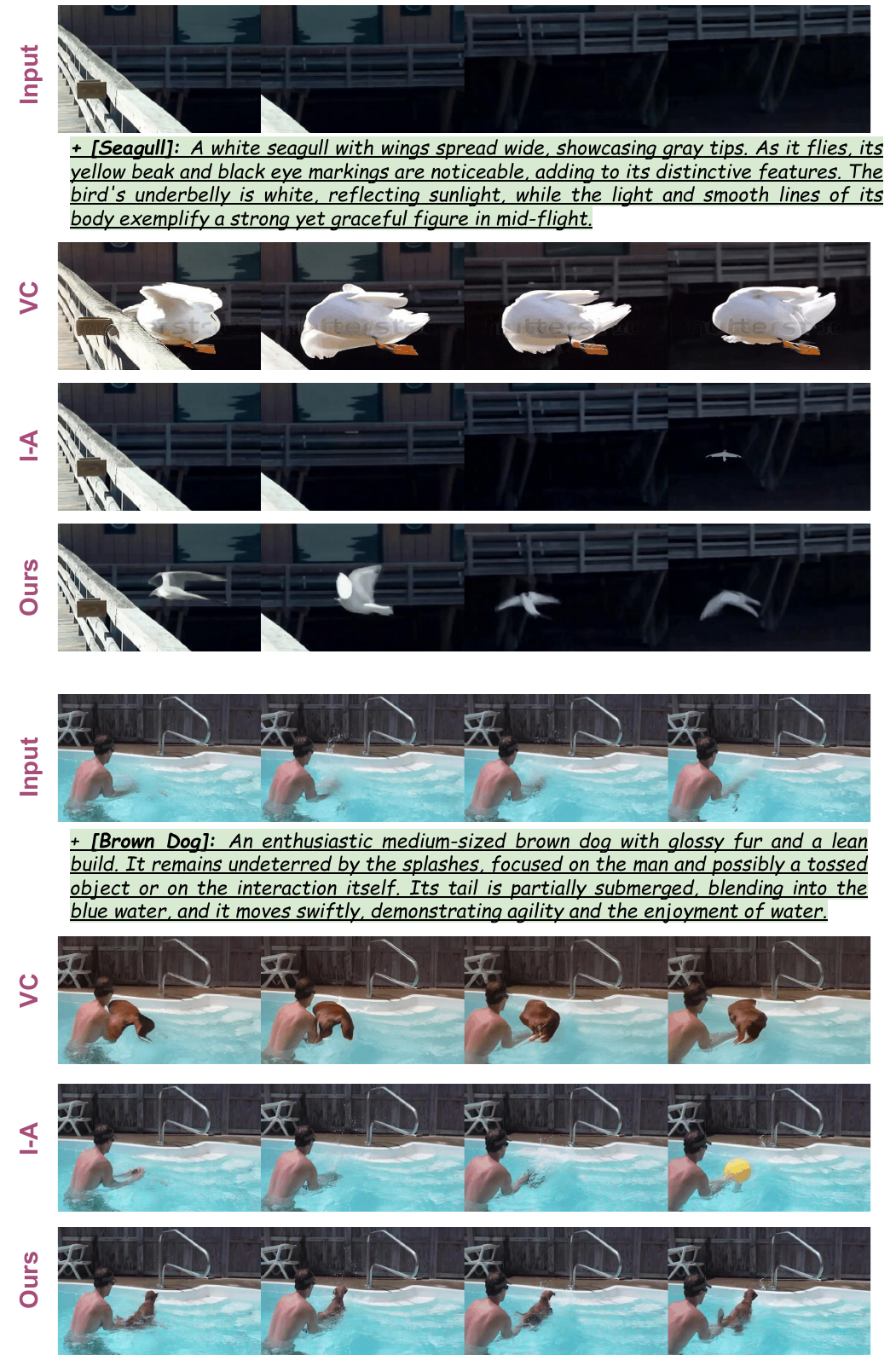}}
    \caption{More visualization of \textbf{adding} video objects}
    \label{fig:add3}
\end{figure*}

\begin{figure*}
    \centering
    {\includegraphics[width=0.9\textwidth]{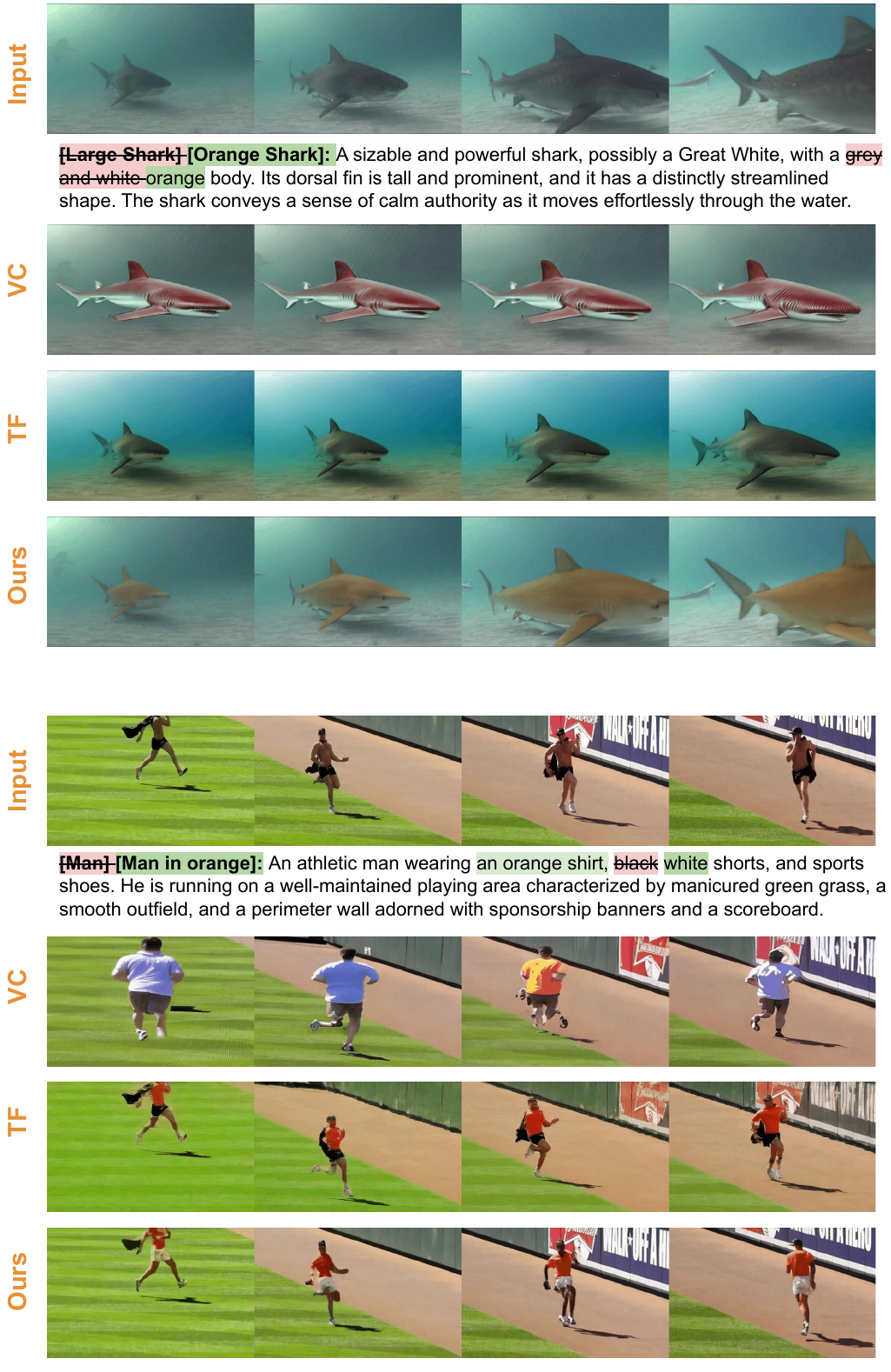}}
    \caption{More visualization of \textbf{editing} video objects}
    \label{fig:editing1}
\end{figure*}

\begin{figure*}
    \centering
    {\includegraphics[width=0.9\textwidth]{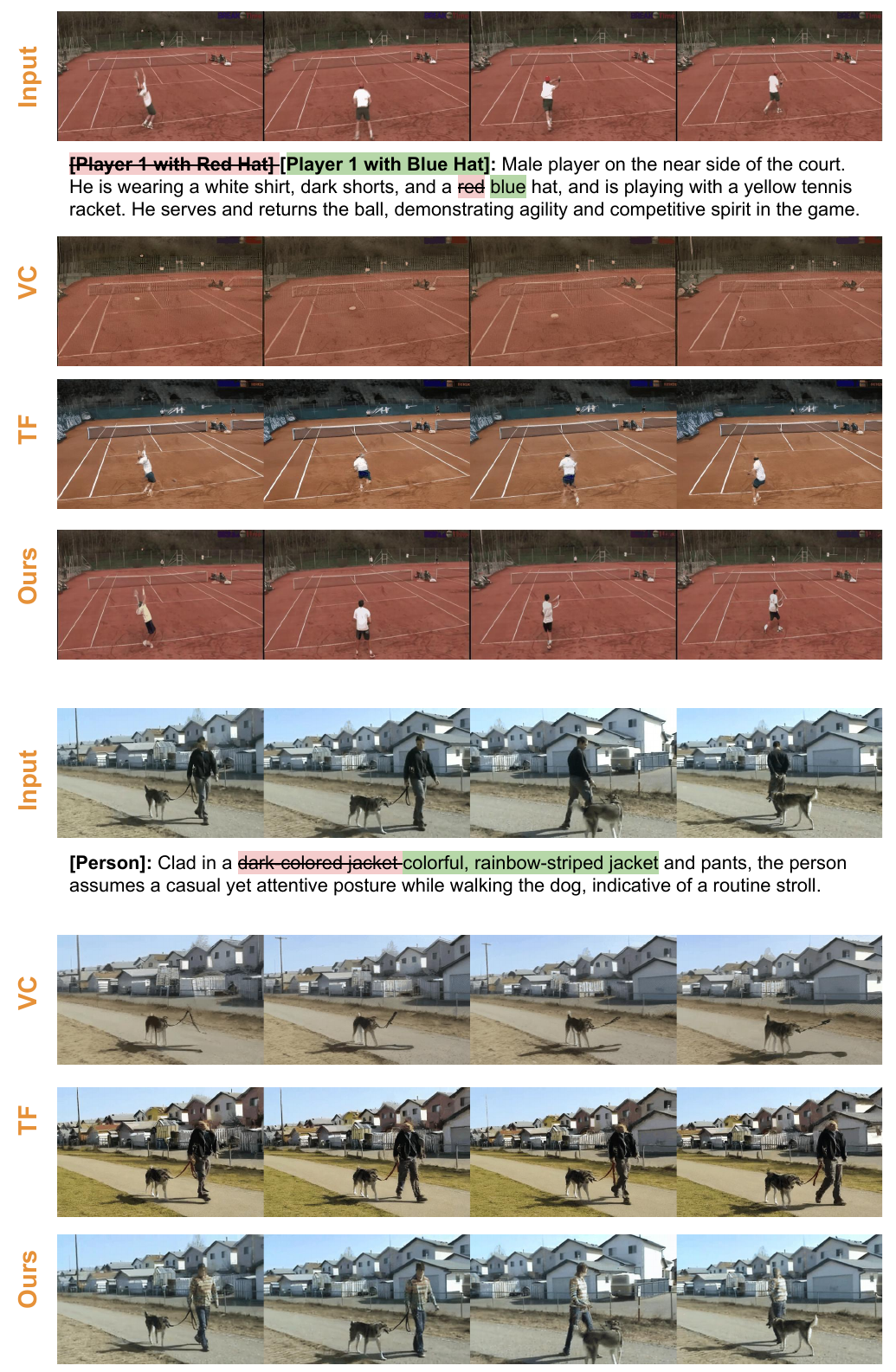}}
    \caption{More visualization of \textbf{editing} video objects}
    \label{fig:editing2}
\end{figure*}

\begin{figure*}
    \centering
    {\includegraphics[width=0.9\textwidth]{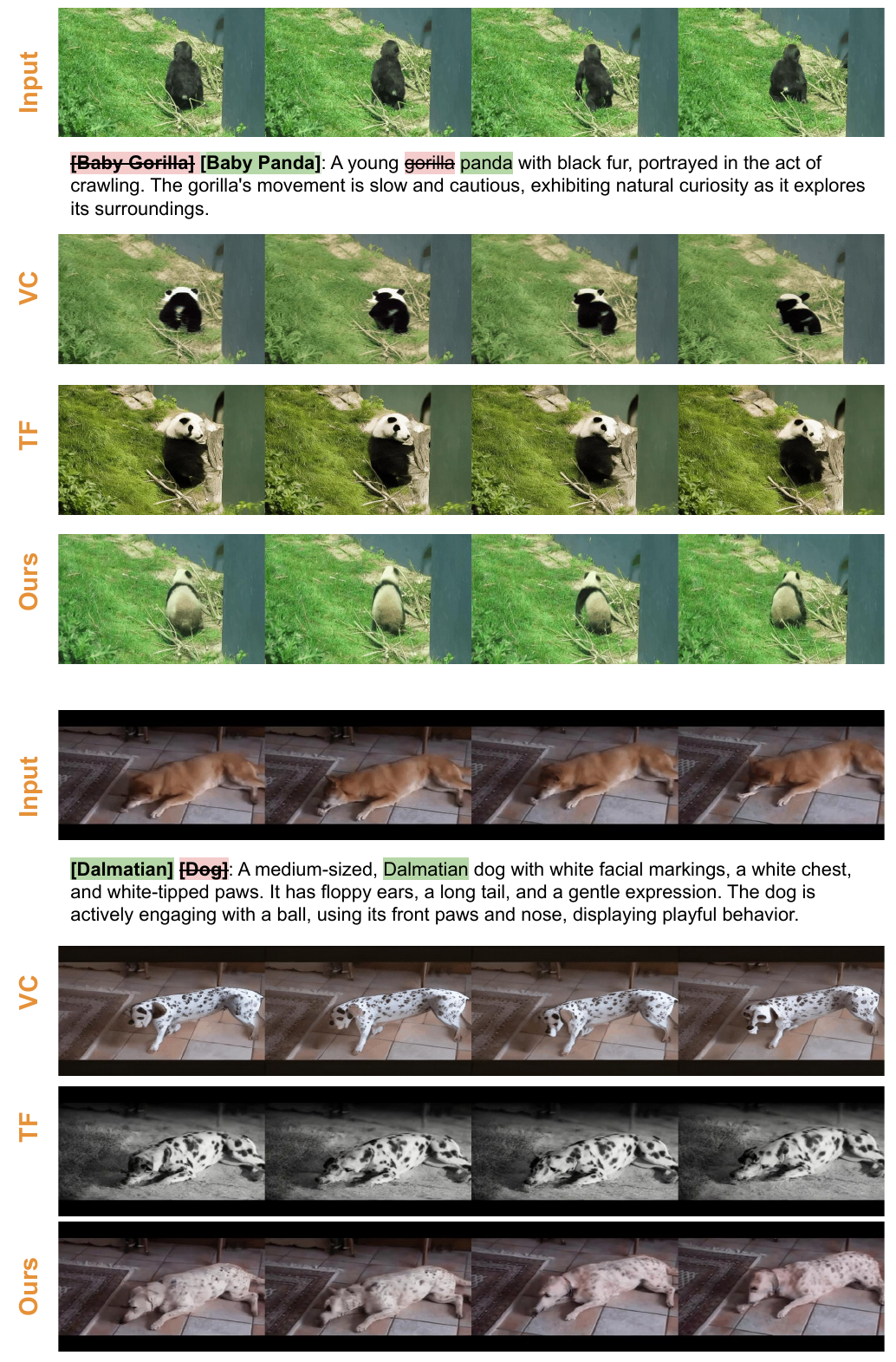}}
    \caption{More visualization of \textbf{editing} video objects}
    \label{fig:editing3}
\end{figure*}

\noindent\textbf{Remove, Add, and Change Object the videos.} We provide more qualitative examples in this Appendix across different types of video content editing (~\cref{fig:sora_example}), including removing (\cref{fig:remove1}, \cref{fig:remove2}, \cref{fig:remove3}), adding (\cref{fig:add1}, \cref{fig:add2}, \cref{fig:add3}), and changing/editing (\cref{fig:editing1}, \cref{fig:editing2}, \cref{fig:editing3}). According to the visualization, our \method generally outperforms other strong baselines on all three subtasks. \method can reflect the updated text description more accurately, thus aiding in user-friendly video editing. For example, our method can accurately change the color of the hat (\textit{red}$\rightarrow$\textit{blue}), which is a very small area in the video, while other methods struggle to meet the requirement.  

\begin{figure*}
    \centering
    {\includegraphics[width=0.9\textwidth]{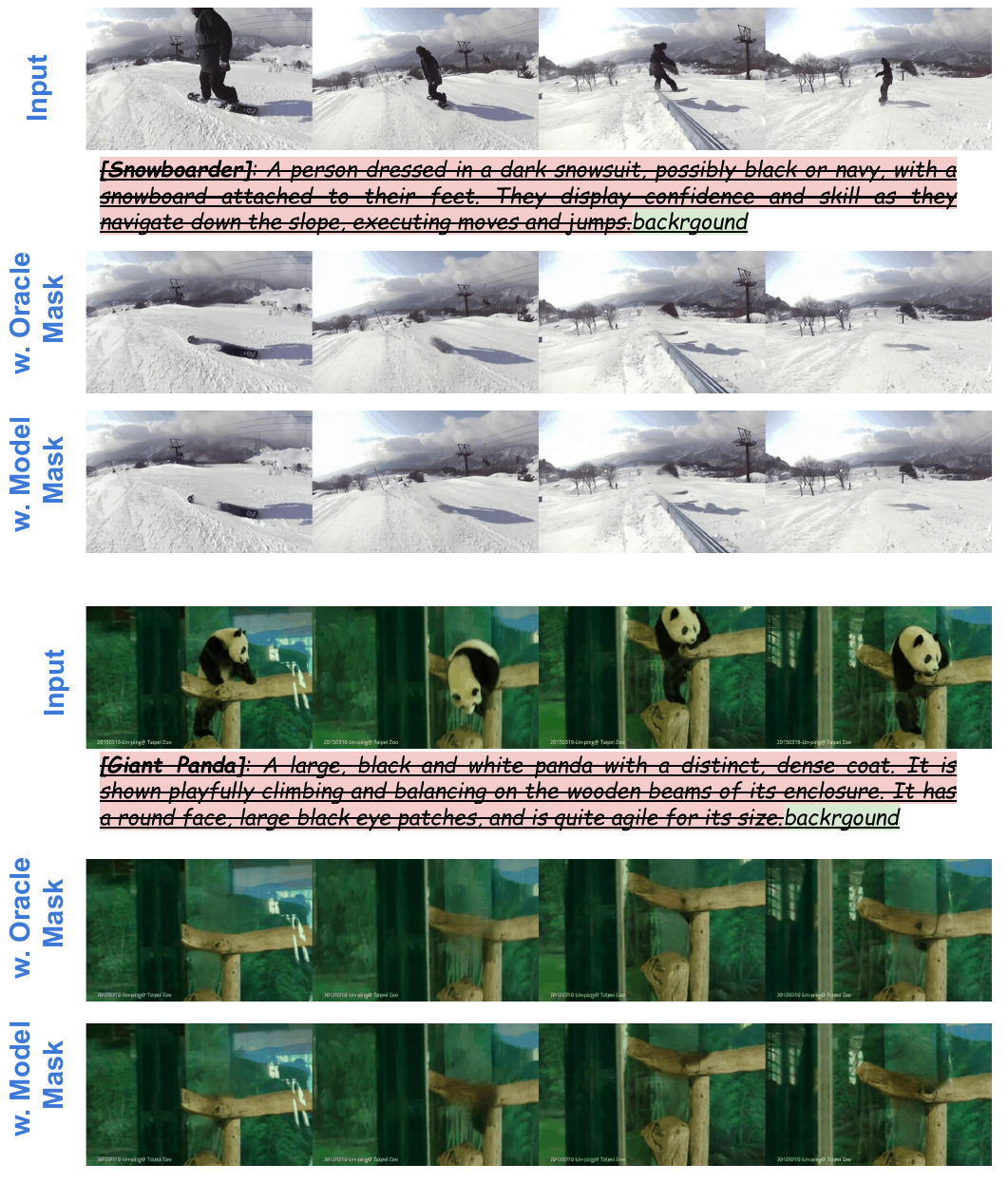}}
    \caption{More visualization of \textbf{removing} video objects}
    \label{fig:remove_ab1}
\end{figure*}

\begin{figure*}
    \centering
    {\includegraphics[width=0.95\textwidth]{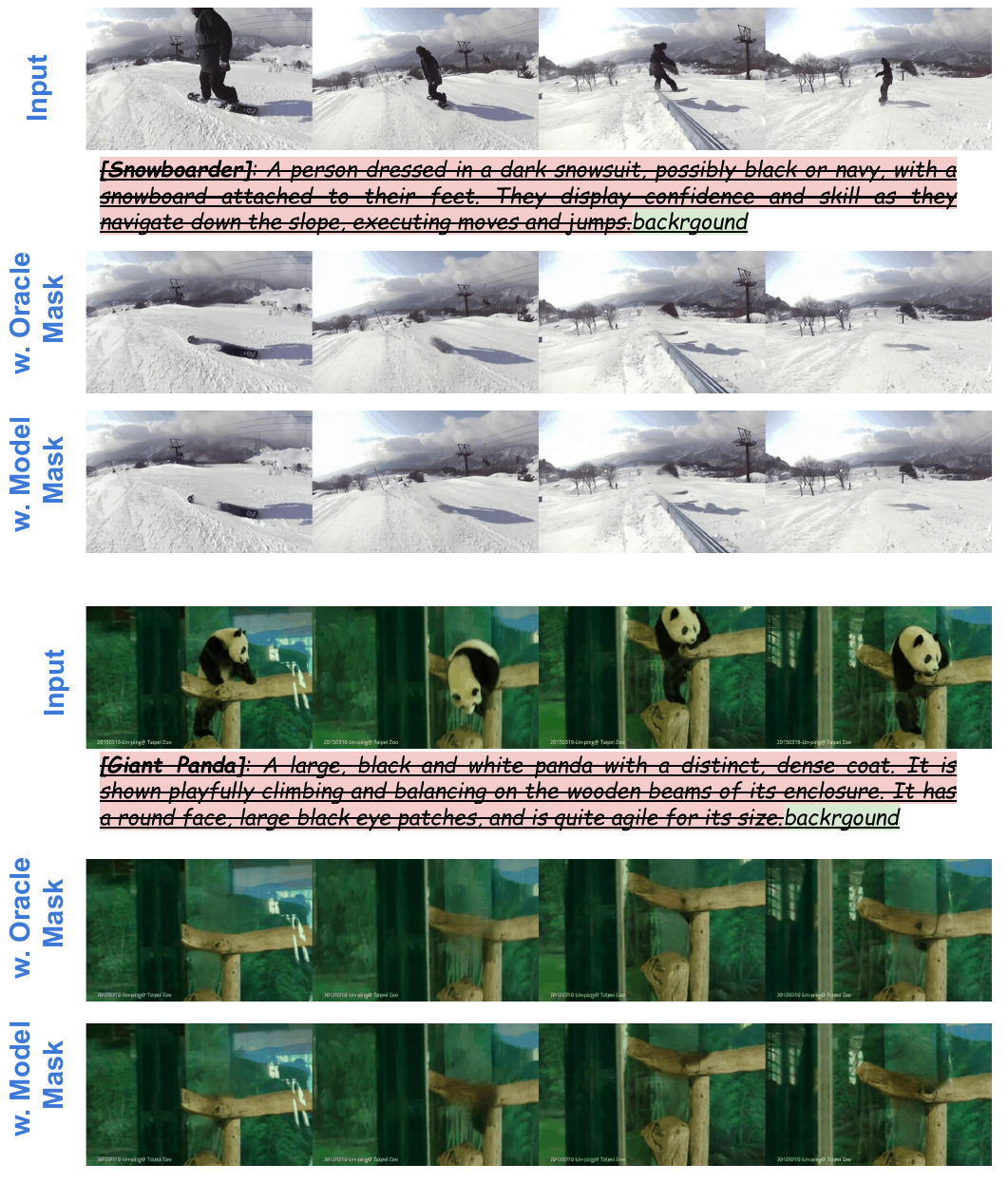}}
    \caption{More visualization of \textbf{removing} video objects}
    \label{fig:remove_ab2}
\end{figure*}

\begin{figure*}
    \centering
    {\includegraphics[width=0.95\textwidth]{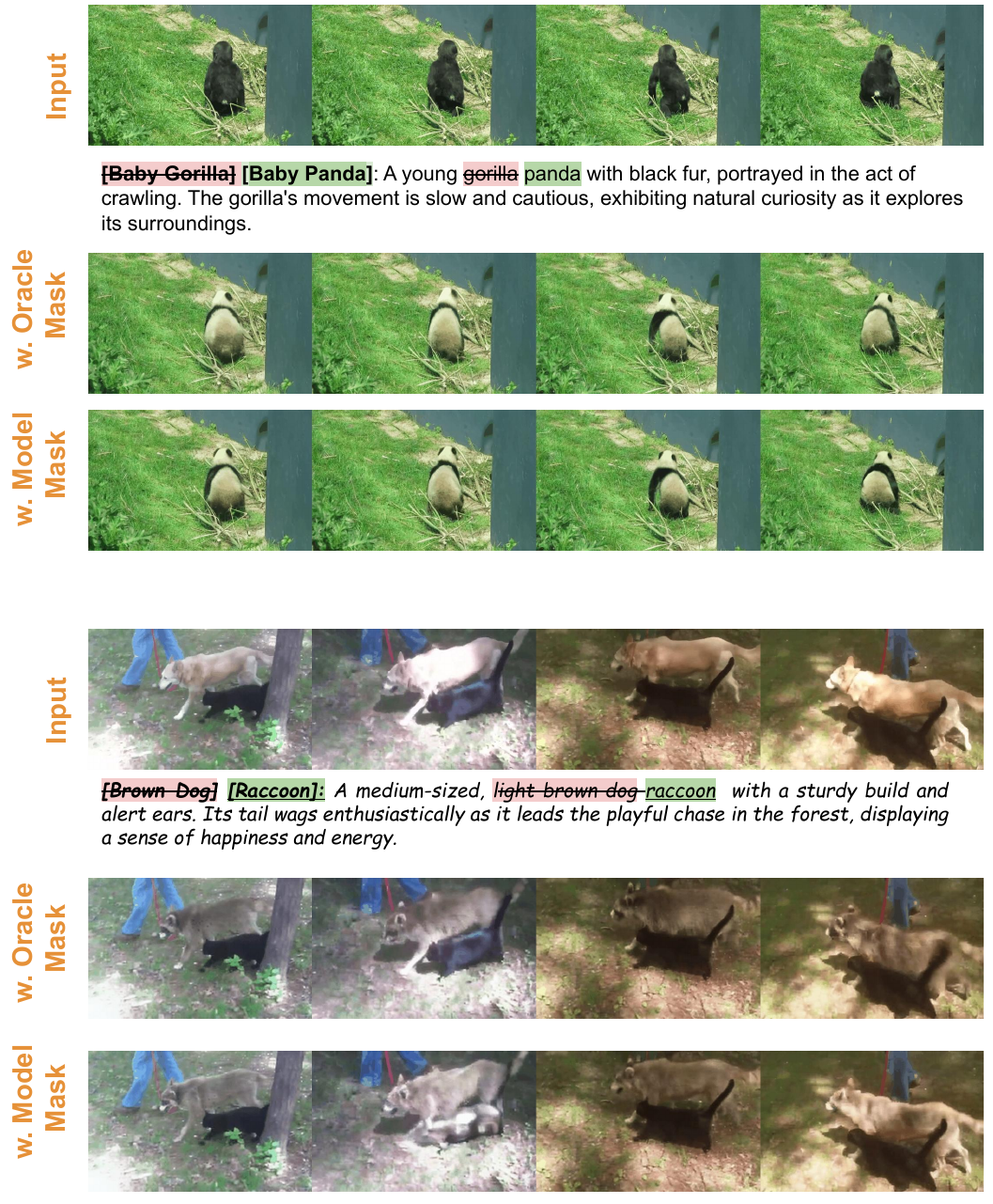}}
    \caption{More visualization of \textbf{editing} video objects}
    \label{fig:editing_ba1}
\end{figure*}

\begin{figure*}
    \centering
    {\includegraphics[width=0.95\textwidth]{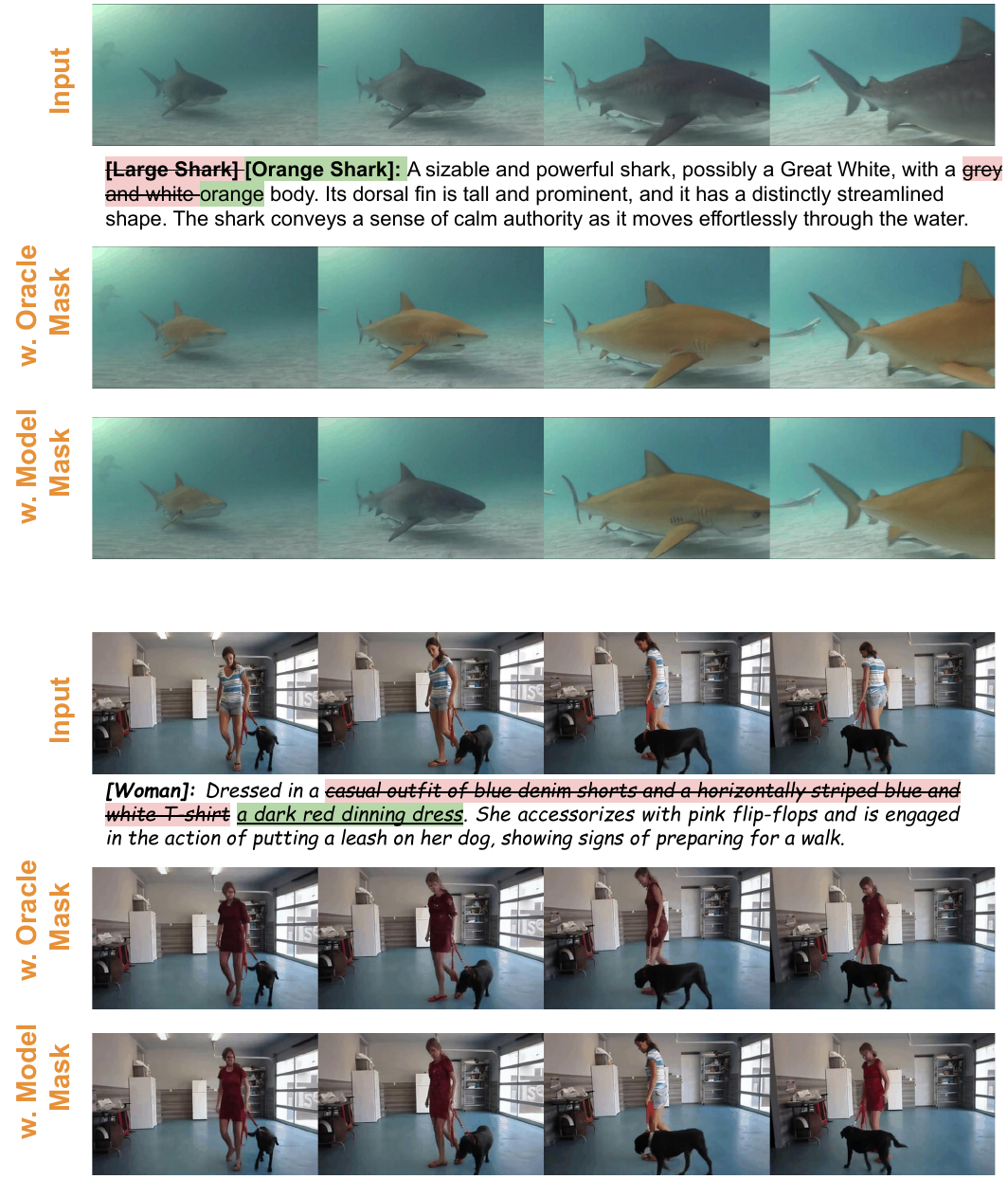}}
    \caption{More visualization of \textbf{editing} video objects}
    \label{fig:editing_ba2}
\end{figure*}

\begin{figure*}
    \centering
    {\includegraphics[width=0.95\textwidth]{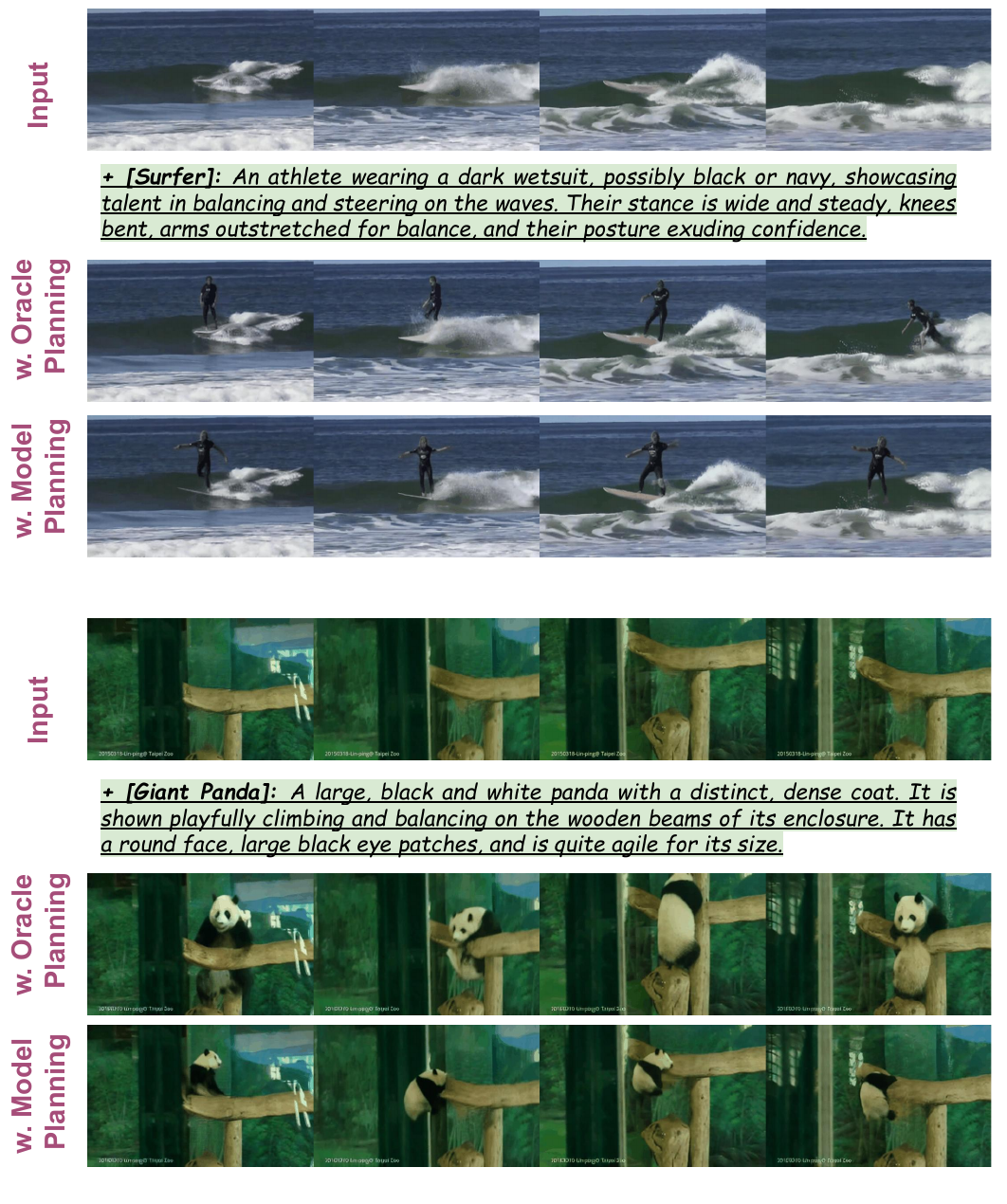}}
    \caption{More visualization of \textbf{adding} video objects}
    \label{fig:add_ab1}
\end{figure*}

\begin{figure*}
    \centering
    {\includegraphics[width=0.95\textwidth]{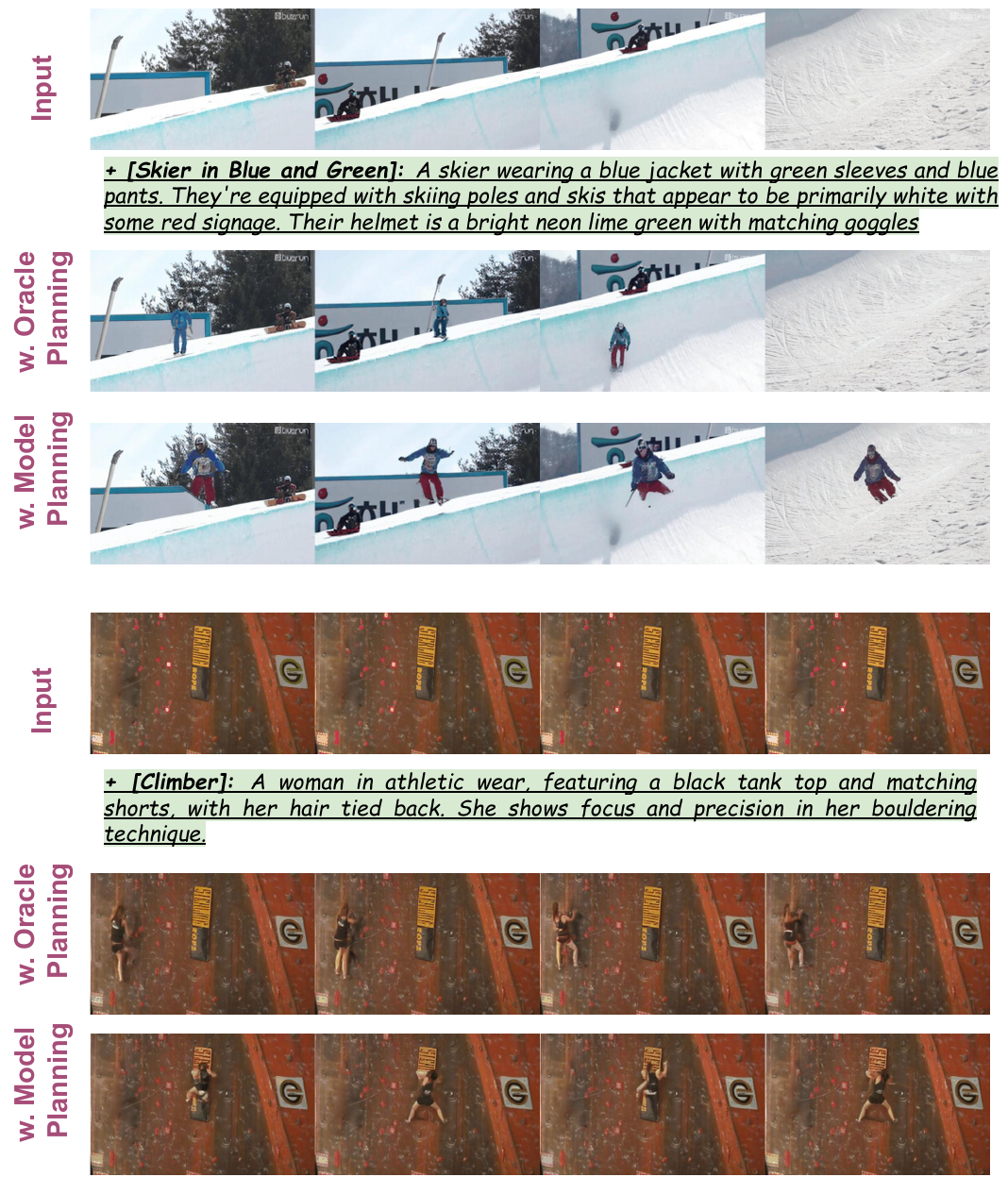}}
    \caption{More visualization of \textbf{adding} video objects}
    \label{fig:add_ab2}
\end{figure*}

\noindent\textbf{Ablation Study Visualization.}
We illustrate extra visualizations for replacing 
orecle mask with grounding\&tracking tools generated ones for video object removal (~\cref{fig:remove_ab1} and ~\cref{fig:remove_ab2}) and changing (~\cref{fig:editing_ba1} and ~\cref{fig:editing_ba2}), as well as replacing oracle object boxes with our model-predicted one (~\cref{fig:add_ab1} and ~\cref{fig:add_ab2}). 
\method shows robust results with LLM planning and off-shelf segmentation tools.
We further show that the failure cases of removing and changing objects mainly come from the missing mask prediction of the video segmentation masks. 






\input{materials/rebuttal_figures}

%% file: materials/apdx_k_means_ablations.tex
\begin{table*}[t]
\caption{\textbf{Ablation of \method} for Video-to-Paragraph Generation on ActivityNet and YouCook2. Metrics are abbreviated: \textbf{M}: \textit{METEOR}, \textbf{B}: \textit{BLEU-4}, \textbf{S}: \textit{SPICE}, \textbf{R}: \textit{ROUGE}. $\bm{v}=1$ indicates the version without superpixel overlap. We highlight the hyperparameter setup used in the main experiment.
}\label{tab:k_means_ablation}
\footnotesize
\centering
\setlength{\tabcolsep}{2mm}
\resizebox{0.75\linewidth}{!}{%
\begin{tabular}{lccccccccccc}
\toprule
\multirow{2}{*}{\textbf{Models}} & \multirow{2}{*}{\textbf{\textit{k}}} & \multirow{2}{*}{\textbf{\textit{v}}} & \multicolumn{4}{c}{\textbf{ActivityNet}}& & \multicolumn{4}{c}{\textbf{YouCook2}}  
\\ \cmidrule(lr){4-7} \cmidrule(lr){9-12}
& \multicolumn{1}{c}{}& \multicolumn{1}{c}{}& \multicolumn{1}{c}{\textbf{S}} & \multicolumn{1}{c}{\textbf{B}} & \multicolumn{1}{c}{\textbf{M}} & \multicolumn{1}{c}{\textbf{R}} & & \multicolumn{1}{c}{\textbf{S}} & \multicolumn{1}{c}{\textbf{B}} & \multicolumn{1}{c}{\textbf{M}} & \multicolumn{1}{c}{\textbf{R}} \\ \midrule
PDVC~\citep{wang2021end} & - & - & - &  2.6 & 10.5 &-&& - & 0.8 & 4.7 &-  \\
Vid2Seq~\citep{yang2023vid2seq} & - & - & 5.4 & - & 7.1 &-  & & 4.0 & - & 4.6 & - \\ 
ZeroTA~\citep{jo2023zero} & - & - & 2.6 & - & 2.7 & - & &  1.6& - & 2.1 & - \\ 
PG-VL~\citep{munasinghe2023pg} & - & -  & 13.6 & 13.9 & 14.2 & 18.1 & & 6.2 & 16.5 & 8.6 & 15.8 \\
\midrule
\multirow{12}{*}{\begin{tabular}{c}
     \textbf{\method}\\\textbf{(Ours)}
\end{tabular}} &   \multirow{4}{*}{20} &  1  &
13.5&13.9& 14.2&18.1& &
6.3&16.9&8.7&15.9\\ 
& &  2  &
13.7& 14.6 &14.4&18.2& &
6.4&17.5&8.7&16.1 \\ 
& & 4 &
13.6&14.3&14.3&18.2& &
6.6&16.2&8.8&16.0\\
&  & 5 & \cellcolor{gg} \textbf{13.8} & \cellcolor{gg}\textbf{15.0} & \cellcolor{gg}\textbf{14.5}& \cellcolor{gg}\textbf{18.4} &&
6.4&17.9&8.8&\textbf{16.2}\\
% &  & 5 & \textbf{13.8} & \textbf{15.0} & \textbf{14.5}& \textbf{18.4} && 6.4&17.9&8.8&\textbf{16.2}\\
\cmidrule(lr){2-12}
&\multirow{4}{*}{25} &  1  &
13.6&14.1&14.3&18.0& &
6.1&16.9&8.6&15.9 \\ 
 &  & 2 &
\textbf{13.8}&14.4&14.3&18.3 &&
6.4&16.3&\textbf{9.0}&16.0\\
&  & 4 &
13.6&14.3&14.3&18.2&&
6.6&17.1&8.9&16.1\\
&  & 6 &
% 13.7&14.5&14.4&18.2& & \textbf{6.9} & 18.0  & \textbf{9.0}& 16.1\\
13.7&14.5&14.4&18.2& & \cellcolor{gg}\textbf{6.9} & \cellcolor{gg}{18.0}  & \cellcolor{gg}\textbf{9.0}& \cellcolor{gg}16.1\\
&  & 10 &
-&-&-&-& & {6.4} & {16.5}  & {8.7}& {16.1}\\
\cmidrule(lr){2-12}
&\multirow{4}{*}{30} &  1  &
13.6&14.1& 14.3& 18.0& &
6.3&16.5&8.8&16.0 \\ 
&  & 2 &
13.7&14.2&14.2 &18.1 &&
6.6&17.1&\textbf{9.0}&\textbf{16.2}\\
&  & 4 &
13.5&14.5&14.3&18.2&&
6.6&\textbf{18.1}&8.8&16.1\\
&  & 6 &
13.6&14.4&14.4&18.2&&
6.4&17.2&8.8&\textbf{16.2}\\
\bottomrule
\end{tabular}}
\end{table*}

%% file: materials/apdx_spixel_init.tex
\begin{table*}[t]
\caption{\textbf{\method variants with different grounding methods} for Video-to-Paragraph Generation on YouCook2.}\label{tab:grounding}
    \centering
    \resizebox{0.9\linewidth}{!}{
    \begin{tabular}{lcccccc}
    \toprule    
    \textbf{Method} & \textbf{Localization} & \textbf{Clustering} & \textbf{SPICE} & \textbf{BLEU-4} & \textbf{METEOR} & \textbf{ROUGE} \\
    \midrule
    PG-VL~\citep{munasinghe2023pg} &-&-& 6.2 & 16.5 & 8.6 & 15.8  \\
    \midrule
    \multirowcell{5}{Ours}&SAM~\citep{kirillov2023segany} &-& 6.4 & 16.9 & 8.7 & 15.9 \\
    &Grounded SAM~\citep{ren2024grounded} &-& 6.5 & 16.5 & 8.7 & \textbf{16.1}  \\
    \cmidrule{2-7}
    &SAM-Track~\citep{cheng2023segment}&k-means& 6.2 & 16.5 & 8.8 & \textbf{16.1}  \\
    &SAM-Track~\citep{cheng2023segment}&overlapping k-means& 6.5 & 17.4 & \textbf{9.0} & \textbf{16.1}  \\
    &\cellcolor{gg}Superpixel&\cellcolor{gg}overlapping k-means& \cellcolor{gg}\textbf{6.9} & \cellcolor{gg}\textbf{18.0} & \cellcolor{gg}\textbf{9.0} & \cellcolor{gg}\textbf{16.1}  \\    
    \bottomrule
    \end{tabular}
    }    
\end{table*}

%% file: materials/rebuttal_figures.tex
% {\includegraphics[width=0.95\textwidth]{figures/vis-edit.pdf}}
% \captionof{figure}{Visualization of text-based video editing. The edited words are marked with \textcolor{BrickRed}{\textbf{Red}}, and the target words are marked with \textcolor{Green}{\textbf{Green}}. 
% \updated{The \method caption is selected from predicted dense captions. 
% We highlight the region of interest with red dashed-line boxes for comparison.}
% More examples are in the Appendix. 
% }
\begin{figure*}[ht]
    \centering
{\includegraphics[width=0.95\textwidth]{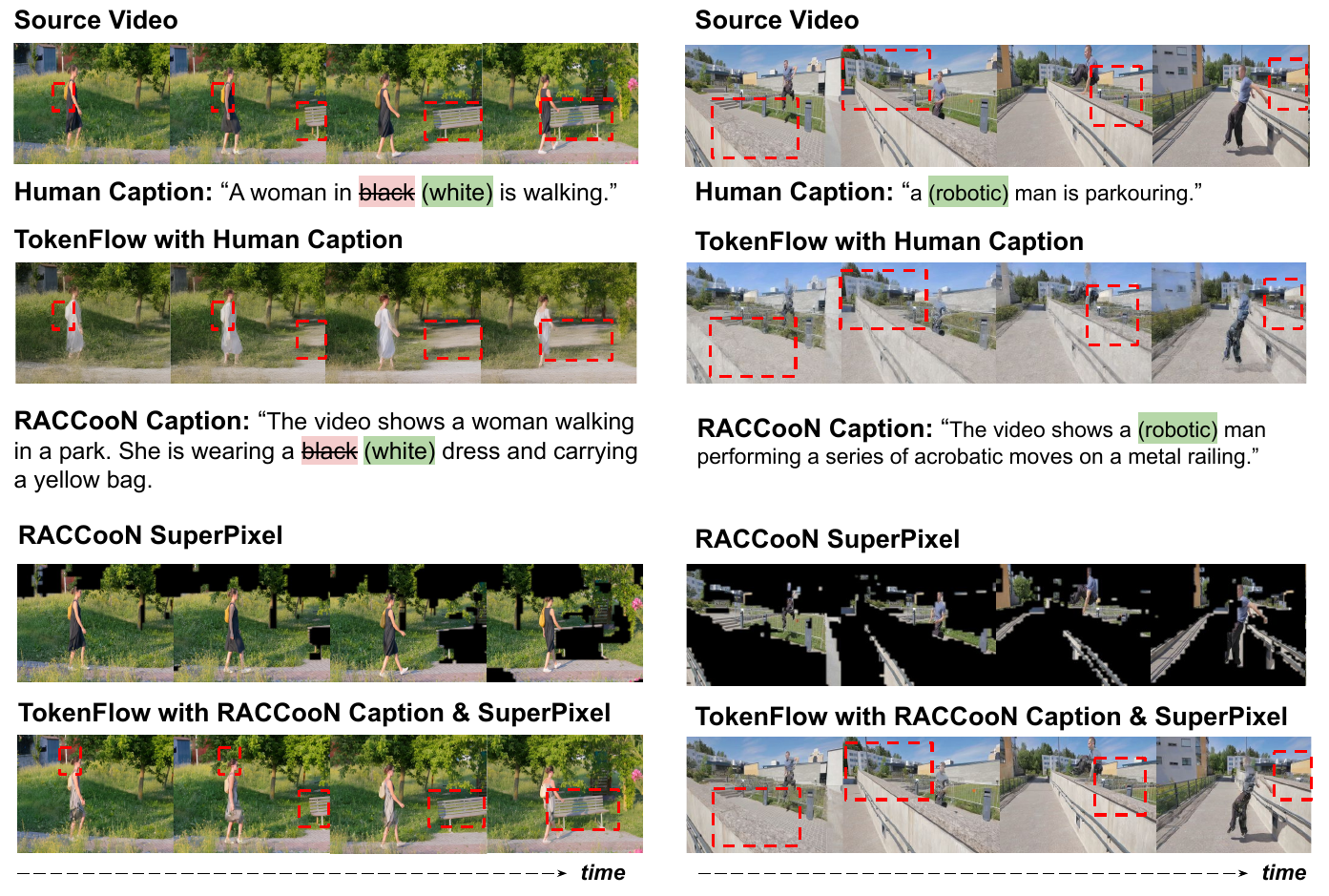}}
    \caption{
    \textbf{Visualization of text-based video editing.} The edited words are marked with \textcolor{BrickRed}{\textbf{Red}}, and the target words are marked with \textcolor{Green}{\textbf{Green}}. The \method caption is selected from predicted dense captions. We highlight the region of interest with red dashed-line boxes for comparison.}
\end{figure*}

\begin{figure*}[ht]
\centering
{\includegraphics[width=0.95\textwidth]{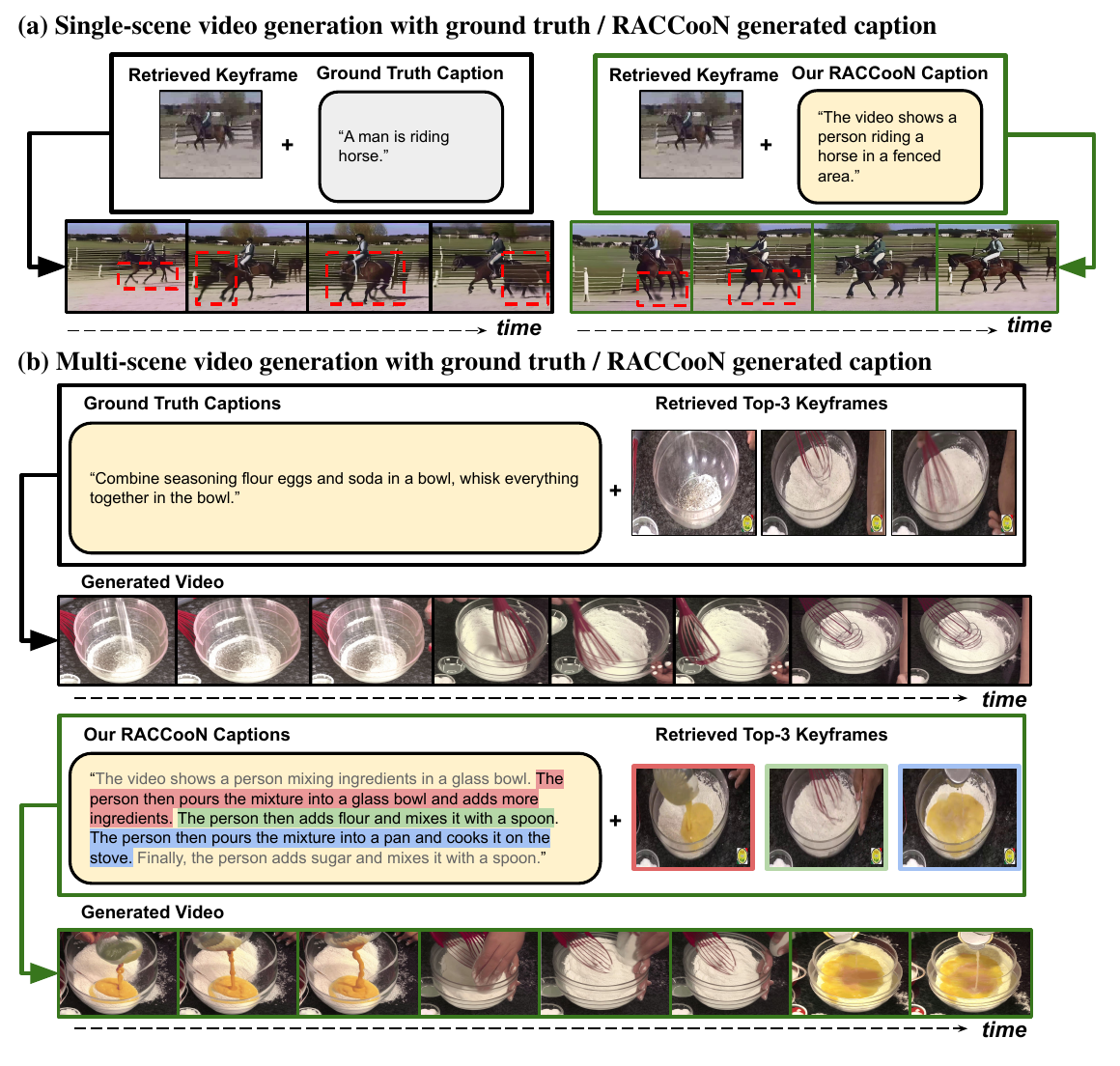}}
\caption{\textbf{Visualization of conditional video generation with VideoCrafter.} Top (a): we compare generation results conditional on different captions and with the \textbf{same} keyframe. Bottom (b): We leverage multiple keyframes retrieved by different captions to generate multi-scene video. We \textcolor{gray}{gray out} captions that are not used for retrieval, and highlight captions used for keyframe retrieval. We highlight the region of distortion with red dashed-line boxes for detailed comparison.}
\end{figure*}

\begin{figure*}[ht]
\centering
{\includegraphics[width=0.95\textwidth]{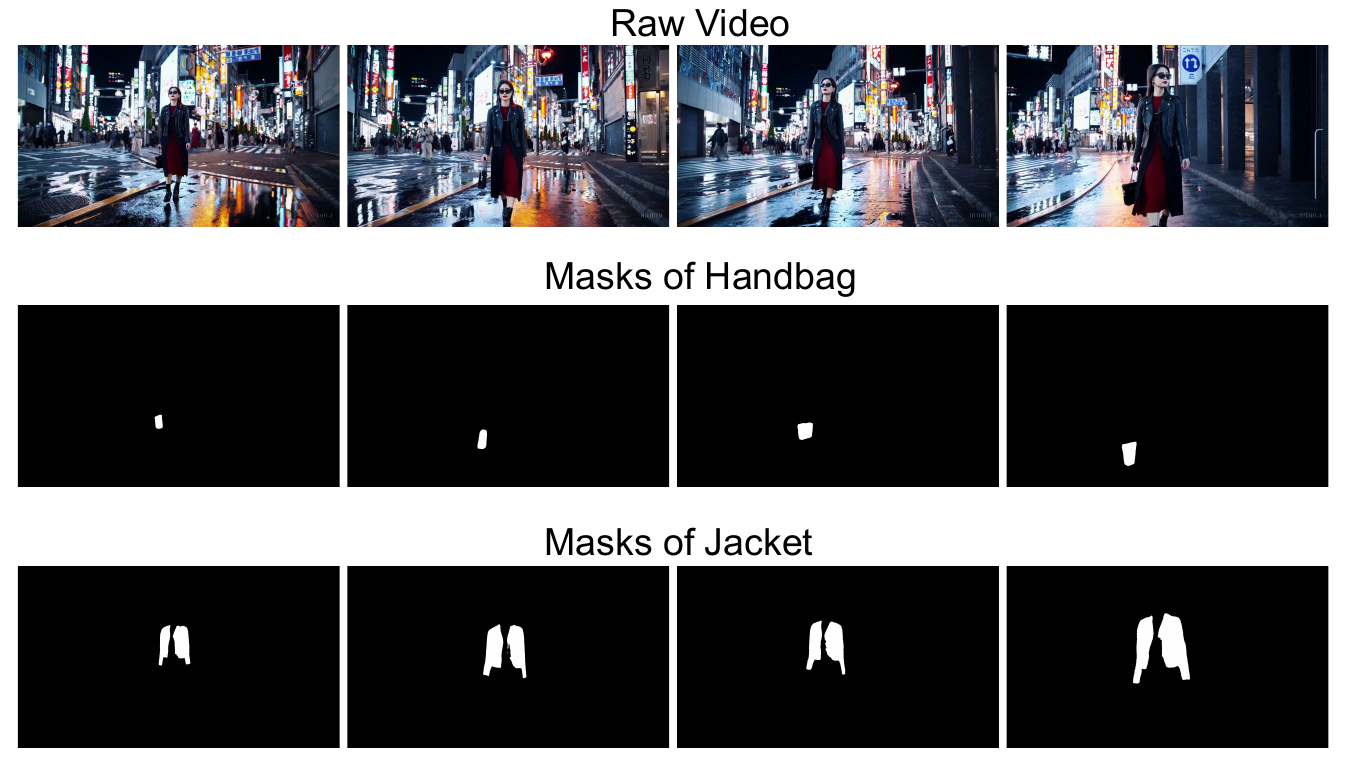}}
\caption{\textbf{Visualization of Masks.}}
\end{figure*}